\documentclass{article}

\usepackage{amsmath}
\usepackage{microtype}
\usepackage{graphicx}
\usepackage{subfigure}
\usepackage{booktabs} % for professional tables
\usepackage{amssymb}
\usepackage{adjustbox} 
\usepackage{multicol} 
\usepackage{mathtools}

\usepackage{enumitem} % make itemize avoid indent

\usepackage{color}

\usepackage{hyperref}
\usepackage{booktabs} % For formal tables

\usepackage[accepted]{icml2024}

\icmltitlerunning{InterLUDE: Interactions between Labeled and Unlabeled Data to Enhance Semi-Supervised Learning}

\begin{document}

\twocolumn[
% \icmltitle{(TEMPORARY)iMatch: Enhancing Semi-Supervised Learning through Deep Interactions between Labeled and Unlabeled Data}
\icmltitle{InterLUDE: Interactions between Labeled and Unlabeled Data \\to Enhance Semi-Supervised Learning}




\icmlsetsymbol{equal}{*}

\begin{icmlauthorlist}
\icmlauthor{Zhe Huang}{equal,tufts}
\icmlauthor{Xiaowei Yu}{equal,uta}
\icmlauthor{Dajiang Zhu}{uta}
\icmlauthor{Michael C. Hughes}{tufts}
\end{icmlauthorlist}

\icmlaffiliation{tufts}{Department of Computer Science, Tufts University, Medford, USA}
\icmlaffiliation{uta}{Department of Computer Science, University of Texas at Arlington, Arlington, USA}

% \icmlcorrespondingauthor{Cieua Vvvvv}{c.vvvvv@googol.com}
% \icmlcorrespondingauthor{Eee Pppp}{ep@eden.co.uk}

\icmlkeywords{Machine Learning, ICML}

\vskip 0.3in
]



\printAffiliationsAndNotice{\icmlEqualContribution} % otherwise use the standard text.

\begin{abstract}
Semi-supervised learning (SSL) seeks to enhance task performance by training on both labeled and unlabeled data. Mainstream SSL image classification methods mostly optimize a loss that additively combines a supervised classification objective with a regularization term derived \emph{solely} from unlabeled data. This formulation neglects the potential for interaction between labeled and unlabeled images. % labeled and unlabeled datasets. 
In this paper, we introduce InterLUDE, a new approach to enhance SSL made of two parts that each benefit from labeled-unlabeled interaction.
The first part, embedding fusion, interpolates between labeled and unlabeled embeddings to improve representation learning.
The second part is a new loss, grounded in the principle of consistency regularization, that aims to minimize discrepancies in the model's predictions between labeled versus unlabeled inputs. 
Experiments on standard closed-set SSL benchmarks and a medical SSL task with an uncurated unlabeled set show clear benefits to our approach.
On the STL-10 dataset with only 40 labels, InterLUDE achieves \textbf{3.2\%} error rate, while the best previous method reports 14.9\%. 

\end{abstract}

\section{Introduction } 
\label{Introduction}
Deep neural networks have revolutionized various fields with their strong performance at supervised tasks. However, their effectiveness often hinges on the availability of large labeled datasets. This requirement presents a significant bottleneck, as labeled data is often scarce and expensive to obtain due to a need for manual annotation by human experts. In contrast, unlabeled data (e.g. images without corresponding labels) may be naturally far more abundant and accessible.
This disparity has led to the growing importance of Semi-Supervised Learning (SSL)~\citep{zhuSemiSupervisedLearningLiterature2005,van2020survey}, a paradigm that jointly trains on small labeled sets and abundant unlabeled data to improve learning when labels are limited. SSL is popular in many learning tasks, such as image classification~\citep{sohn2020fixmatch}, object detection~\citep{xu2021end,li2022rethinking}, and segmentation~\citep{chen2021semi,yang2023revisiting}. 

This paper focuses on SSL for image classification. Over the years, numerous SSL paradigms have been proposed ~\citep{blum1998combining,min2020mutually, kingma2014semi,kumar2017semi,nalisnick2019hybrid,liu2019deep,iscen2019label}.
The current prevailing paradigm trains deep neural classifiers to jointly optimize a supervised classification objective and a regularization term derived \emph{solely} from the unlabeled data. 
%due to its simplicity and effectiveness~\citep{oliver2018realistic,rizve2021defense}. 
Examples of this paradigm that report state-of-the-art results are numerous~\citep{laine2016temporal,tarvainen2017mean,miyato2018virtual,sohn2020fixmatch,xu2021dash,zhang2021flexmatch,wang2022freematch,chen2023softmatch}.

Despite these advancements, a critical gap persists in how these SSL algorithms engage with both labeled and unlabeled data. Notably, there is a \emph{disconnect} between the two data types throughout training~\citep{huang2023flatmatch}. We contend that this lack of deeper interaction fails to fully harness the potential of unlabeled data.

\let\thefootnote\relax\footnotetext{Open-source code: upon acceptance}

In response to this challenge, we introduce InterLUDE, a novel SSL algorithm that facilitates direct interaction between labeled and unlabeled data, substantially enhancing the learning outcomes. Our contributions are summarized as follows: 
% In response to this challenge, we introduce iMatch, a novel SSL algorithm that more effectively utilizes training signals from unlabeled data by fostering direct interplay between labeled and unlabeled data, thereby enhancing the overall learning process.

\begin{itemize}[leftmargin=*]
    \item First, we introduce embedding fusion (Sec. \ref{EmbeddingFusion}), a key part of the InterLUDE training process that interpolates between labeled and unlabeled embedding vectors to improve representation learning. Ablations in Fig.~\ref{fig:CNN_Ablation_FusionStrategy_CIFAR10_40labels} suggest labeled-unlabeled interaction is specifically useful here.

    \item Second, we introduce the cross-instance delta consistency loss (Sec. \ref{RelativeLoss}),
    a new loss that makes changes (deltas) to a model's predictions similar \emph{across} the labeled and unlabeled inputs under the same augmentation change. 
     
    % regulate model behavior \emph{across} labeled and unlabeled samples under the same augmentation changes. Here, labeled and unlabeled data interact as a form of consistency regularization.

    % that aims to make changes in model prediction consistent \emph{across} labeled and unlabeled data under the same changes in data augmentations. Here, labeled and unlabeled data interact as a form of consistency regularization.
    
    % This loss aims to make changes to a model's predicted class probabilities across augmentations similar for the labeled and unlabeled inputs. Here, labeled and unlabeled samples interact as a form of consistency regularization.

    \item Our final contribution is that our experiments cover both standard benchmarks (Sec.~\ref{Experiments}) as well as a more realistic ``open-set'' medical task (Sec.~\ref{Experiments_OpenSSL}). Typical recent work in SSL (e.g. ~\citet{wang2022freematch}) mostly assumes that labeled and unlabeled data come from the same distribution.
    However, in the intended real-world applications of SSL, unlabeled data will be collected automatically at scale for convenience, and thus can differ from the labeled set~\citep{oliver2018realistic}. When unlabeled images contain extra classes beyond the labeled set, this is called ``open-set'' SSL~\citep{yu2020multi,guo2020safe}.
    To improve SSL evaluation practices, we evaluate on the open-set Heart2Heart benchmark proposed by \citet{huang2023fix}, as well as standard datasets (CIFAR and STL-10).

%    Standard SSL . However, this assumption often does not hold in practice, as unlabeled datasets are typically collected automatically at scale, leading to an uncurated and diverse nature ~\citep{oliver2018realistic}. This more realistic scenario is often referred to as ``open-set'' SSL ~\citep{yu2020multi,chen2020semi,guo2020safe}, in contrast to the traditional ``closed-set'' SSL. While the primary focus of our method is on the classic SSL setting as in~\citet{wang2022freematch}, our later evaluations will explore an ``open-set'' medical SSL benchmark introduced in~\citet{huang2023fix}.

\end{itemize}

Ultimately, our experiments encompass \textbf{six diverse datasets} and \textbf{two architecture families}, including Convolutional Neural Networks (CNNs) and Vision Transformers (ViTs). 
Across scenarios, we find InterLUDE is effective at both closed-set natural image tasks and open-set medical tasks. This latter finding is exciting because InterLUDE was not deliberately designed to handle open-set unlabeled data.
We hope this work on InterLUDE opens a new avenue for future SSL research, emphasizing the extraction of training signals from the \textit{interplay} between labeled and unlabeled data, rather than processing each modality in isolation.

\section{Background and Related Work}
\label{RelatedWork}
%\subsection{Semi-supervised learning}
\textbf{Semi-supervised learning.}
In semi-supervised image classification problems, we are given a labeled dataset $\mathcal{D}^L$ of image-label pairs $(x_{l}, y)$, and a much larger unlabeled dataset $\mathcal{D}^U$ containing only images $x_{u}$ (i.e., $|\mathcal{D}^U| \gg |\mathcal{D}^L|$). Given both data sources, our goal is to train a classifier that maps each image to a probability vector in the $C$-dimensional simplex $\Delta^C$ representing a distribution over $C$ class labels. 
Comprehensive reviews of SSL can be found in~\citet{zhuSemiSupervisedLearningLiterature2005,van2020survey}. 

In this paper, we focus on the current dominant paradigm, which trains the weights $\theta$ of a deep neural network $f$ to minimize a two-task additive loss
\begin{align} % Use \! to reduce whitespace between symbols
\resizebox{.91\hsize}{!}{$
\min_{\theta} ~ \!
	\textstyle \sum_{x_{l},y \in \mathcal{D}^L} \!
		\ell^L( y, f_{\theta}(x_{l}) )	
	+ 
	\lambda \!
	\sum_{x_{u} \in \mathcal{D}^U} \!
		\ell^U(f_{\theta}(x_{u}))
	\label{eq:standard_ssl_objective}
$}
\end{align}
The loss $\ell^L$, computed \emph{solely} from the labeled set, is most often the standard cross-entropy loss used in supervised classifiers.
The loss $\ell^U$ is a method-specific loss that is typically based \emph{solely} on the unlabeled set. 

Popular approaches under this paradigm include \textit{Pseudo-labeling} that encourages the classifier to assign high probability to confidently-predicted labels for unlabeled images~\citep{lee2013pseudo,arazo2020pseudo,cascante2021curriculum} and \textit{Consistency regularization}~\citep{sajjadi2016regularization,tarvainen2017mean} that enforces consistent model outputs for the same unlabeled image under different transformations.
Recent \textit{hybrid} methods~\citep{berthelot2019mixmatch,sohn2020fixmatch,wang2022freematch} combines several techniques.
While successful, these approaches lack direct\footnote{Indirect interactions do exist, such as 
due to batch normalization~\citep{zhao2022out} or how training combines the two losses.} labeled-unlabeled interaction. We argue this limitation prevents fully harnessing the potential of SSL.

\textbf{Direct labeled-unlabeled interaction in past methods.}
Several past works do engineer some direct interactions.
MixMatch~\citep{berthelot2019mixmatch} allows labeled-unlabeled interaction within their augmentation procedure. Their mixup-style interpolation can randomly choose to blend labeled and unlabeled images and corresponding labels or pseud-labels.
However, MixMatch's stated goal is to ``unify dominant approaches'' to SSL; they do not specifically argue for labeled-unlabeled interaction.

SimMatch~\citep{zheng2022simmatch} encourages weakly and strongly augmented unlabeled samples to maintain the same similarity with labeled embeddings in a memory bank. 
%This is further intensified through unfolding and aggregation operation. %% MCH cut as a detail that did not add much
SimMatchV2's~\citep{zheng2023simmatchv2} edge-node consistency term uses labeled samples to calibrate the pseudo-labels of weakly augmented unlabeled samples.
%, thus creating a direct link between the two datasets. 
Compared to SimMatch, our InterLUDE eliminates the need for additional memory banks and achieves notably better performance.

The concurrent work most similar in spirit to ours is FlatMatch 
~\citep{huang2023flatmatch}. 
%They highlight the disconnect between labeled and unlabeled data in SSL, pointing out that consistency is often only enforced ``instance-wise'' (over different transformations of the same image instance). 
To address the disconnect between labeled and unlabeled data, FlatMatch employs sharpness-aware minimization (SAM)~\citep{foret2020sharpness,kwon2021asam,liu2022towards} to ensure that predictions on unlabeled samples are consistent across both the regular model and a worst-case model, generated by parameter perturbations that maximize empirical risk on labeled data. This approach, however, increase the computation cost due to the additional back-propagation required, a challenge partially alleviated by an approximation strategy they call FlatMatch-e. In contrast, our InterLUDE avoids any additional back-propagation or lossy approximation, runs much faster ($\sim$8x faster in wall-time comparison, see App.~\ref{sec:WallTimeComparison_TMED2})), and achieves better performance (see Table.~\ref{wideresnet-table} \& \ref{tab:Heart2Heart}).

\textbf{Data augmentation in SSL.}
% \subsection{Data Augmentation in SSL}
% % Adversarial Perturbation~\citep{miyato2018virtual}, 
Data augmentation is an integral part of many modern SSL algorithms~\citep{laine2016temporal,xie2020unsupervised}. Common techniques include random flip and crop~\citep{krizhevsky2012imagenet}, MixUp~\citep{zhang2017mixup}, CutOut~\citep{devries2017improved}, AutoAugment~\citep{cubuk2018autoaugment}, and RandAugment~\citep{cubuk2020randaugment}. In SSL, augmentation has primarily been confined to the image space; perturbings of embeddings remain underexplored. %%~\citep{kuo2020featmatch}.
FeatMatch~\citep{kuo2020featmatch} is a rare example of SSL that perturbs embedding vectors (not images), using learned class prototypes. In a similar vein, our embedding fusion directly manipulates the embedding space. Yet, our approach diverges significantly in procedure (avoiding class prototypes) and yields better performance (see Table~\ref{wideresnet-table}). 
\begin{figure*}[htb]
  \centering
 \includegraphics[width=17cm]{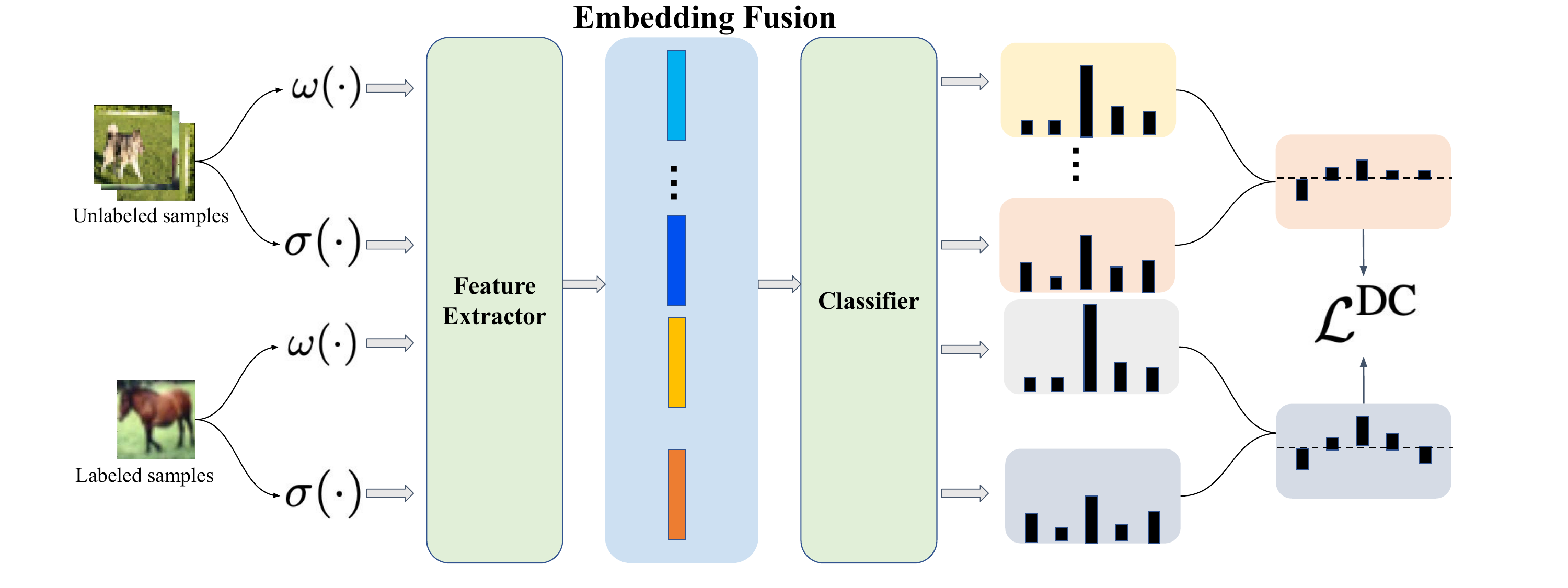}
  % \includegraphics[width=17cm]{figs/CentralFigure_20240110.pdf}

%  \vspace{2.0cm}
  \caption{InterLUDE Framework. The labeled and unlabeled data are individually subjected to both weak and strong augmentations, followed by the backbone extracting feature embeddings. Embedding fusion then perturbs the embeddings as shown in Fig~\ref{embeddingfusion}.
  The delta consistency loss is employed to regulate model behavior on labeled and unlabeled samples under the same augmentation changes.}
\label{methodfig}
\end{figure*}

% \caption{iMatch Framework. The labeled and unlabeled data are individually subjected to both weak and strong augmentations. Subsequently, the backbone extracts embeddings from these augmentated samples. The process of embedding fusion involves employing effective interaction strategies to connect the latent embeddings derived from both labeled and unlabeled samples. The relative loss function is utilized to impose regularization on the relative differences observed in predictions resulting from the weak and strong augmentations applied to the unlabeled and labeled data samples. }

% \section{Method: Enhancing SSL by Bridging Labeled and Unlabeled Data}
\section{Method}
\label{Method}
\textbf{Problem setup.}
Denote the overall classifier $f$ with weights $\theta$ as a composition of two functions: $f = h \circ g$. Neural network $g$ maps an input image $x_i$ to an embedding representation $z_i \in \mathbb{R}^D$ with $D$ dimensions. Practitioners can set $D$ via architectural choices.
Neural network $h$ then maps the $D$-dimensional embedding to a probability vector in the C-dimensional simplex $p_i \in \Delta^C$ over the $C$ classes. In practice, we set $g$ as all layers from input to the second-last layer for a given $f$, following evidence from~\citet{Yu2023PN} suggesting that deeper layers generally yield better performance.

We train $f$ using stochastic gradient descent, where each minibatch 
contains $B$ images from the labeled set and $\mu \cdot B$ images from the unlabeled set. We fix $\mu=7$ following past work~\citep{sohn2020fixmatch,wang2022freematch}.
Thus, each batch contains in total $R = (1 + \mu)B$ distinct images.

% In practice, given a fixed architecture for $f$, we set the embedding network $g$ as all layers from input to the second-last layer, following evidence from~\citet{Yu2023PN} suggesting that deeper layers generally lead to better performance.

%in CNNs and ViTs. In CNNs, $D$ includes $hwc$, where $h$, $w$, and $c$ are the height, width, and channel of the feature map. In ViTs, $D$ includes $hw$, where $h$ is the number of patches, and $w$ is the embedding dimension of a specific patch.

% We wish to train this classifier $f$ in a semi-supervised manner, given access to both labeled dataset $\mathcal{D}^L$ and unlabeled data $\mathcal{D}^U$. We pursue stochastic gradient descent training, sampling small batches of data deliberately constructed 

% We train $f$ in a semi-supervised manner using stochastic gradient descent. Each minibatch 
% contain $B$ images from the labeled set and $\mu \cdot B$ images from the unlabeled set. We fix $\mu=7$ following the convention~\citep{sohn2020fixmatch,zheng2023simmatchv2,wang2022freematch}.
% Thus, each batch contains total $R = (1 + \mu)B$ distinct examples.

We apply both \emph{weak} (random flip and crop) and \emph{strong} augmentations 
 (RandAugment as in FixMatch~\citep{sohn2020fixmatch}) to each image.
Let $\Omega$ and $\Sigma$ denote \emph{sets} of possible weak and strong augmentations. 
We can draw specific realizations $\omega$ from $\Omega$ or $\sigma$ from $\Sigma$. Each realization defines a specific transformation (e.g. rotate by 15 degrees). After augmentation, each batch contains $Q=2R$ samples.

\textbf{Overview of InterLUDE.}
Our proposed InterLUDE algorithm comprises two main components. First, an \emph{embedding fusion} strategy that improves representation quality. Second, a new loss term called \emph{cross-instance delta consistency} that makes changes (deltas)
to a model’s predictions similar \emph{across} the labeled and
unlabeled inputs under the same augmentation change ($\omega$ vs $\sigma$). These two components each promote labeled-unlabeled (LU) interaction and ultimately work in synergy to improve model performance.
Fig. \ref{methodfig} illustrates the InterLUDE framework. Alg.~\ref{alg:iMatch}
provides pseudocode. Details on each component are introduced below. 

% encourages the same augmentations ($\omega$ and $\sigma$) to have a similar impact on predicted probabilities \emph{across} labeled and unlabeled examples.
% These two components each promote labeled-unlabeled (LU) interaction and ultimately work in synergy to improve classifier performance.
% Fig. \ref{methodfig} illustrates the InterLU framework. Alg.~\ref{alg:iMatch}
% provides pseudocode. Details on each component are introduced below. 

\subsection{Embedding Fusion: \emph{Better embedding via interaction}}
\label{EmbeddingFusion}

Emerging evidence suggests that \textbf{proactively perturbing} the embedding space can yield significant performance gains across various learning tasks, such as supervised image classification~\citep{verma2019manifold,Yu2023PN}, domain adaptation~\citep{Yu2023PN}, Natural Language Processing (NLP)~\citep{Pereira2021,khan2023art} and fine-tuning large language models~\citep{Jain2023PN}. Our work explores deliberate perturbation of embeddings via labeled-unlabeled interaction to improve semi-supervised learning.

\begin{figure}[ht]
  \centering
 \includegraphics[width=7 cm]{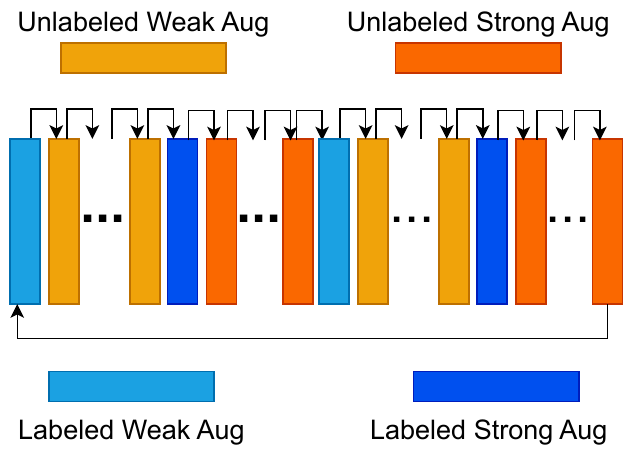}
 % \vspace{2.0cm}
  \caption{Illustration of Embedding Fusion with Batch Size of 2. This operation perturbe each embedding slightly by its immediate previous neighbor embedding. Showing the flattened embeddings for clear visualization. \iffalse Embedding Fusion perturbs the feature embeddings by circularly combining them for samples in the minibatch.\fi}
\label{embeddingfusion}
\end{figure}

Given the batch of $Q$ augmented labeled and unlabeled samples, we map via the network $g$ to an array of embeddings $Z \in \mathbb{R}^{Q\times D}$. Key to our embedding fusion strategy is a specific \textbf{interdigitated layout} arrangement of $Z$ (illustrated in
Fig.~\ref{embeddingfusion}).
By construction, each set of $1+\mu$ adjacent rows in $Z$ (labeled embedding followed by $\mu$ unlabeled embeddings) is created using the same specific augmentation $\omega$ or $\sigma$ (see Alg.\ref{alg:iMatch}).

% To begin, our embedding fusion requires that we obtain weak and strong augmentations of each of the $R$ input images in the batch, yielding $Q = 2R$ total images. We stack them in the interdigitated layout shown in Fig. TODO, yielding an array $X$ of shape $Q \times P$, where $P$ is the input dimension. 
% Mapping each image in order through our embedding network $g$, we obtain an array $Z$ of shape $Q \times D$.

\textbf{General Framework for Embedding Fusion.}
% For any $Z \in \mathbb{R}^{Q \times D}$, we imagine a deterministic fusion transformation parameterized by a matrix $A \in \mathbb{R}^{Q \times Q}$:
For any $Z$, we imagine a deterministic fusion transformation parameterized by a matrix $A \in \mathbb{R}^{Q \times Q}$:
\begin{align}
    Z' \gets (I + A) Z
\end{align}
where $I$ is the identity matrix. Each row of $Z'$ is a linear combination of the rows of $Z$, thus \emph{fusing} together original embeddings. Predicted class probabilities for each image can then be obtained via feeding each row of $Z'$ into the classification head $h$. This construction is inspired by \citet{Yu2023PN}, who pursue embedding fusion for the supervised (not semi-supervised) case.

We impose three constraints on matrix A~\citep{Yu2023PN}:
% imagine three desirable properties that a matrix $A$ must obey for this kind of design to be beneficial:
\begin{align}
\label{eq:desiderata}
(i)~ & \text{rank}(I+A) = Q 
\\ \notag 
(ii)~ & \left [ I+A\right ]_{ii} > \left [ I+A \right ]_{ij}, i \ne j 
\\ \notag
(iii)~ & \left \| \left [ I+A\right ]_{i}   \right \|_{1} = 1
\end{align}
\emph{The first constraint}, the full rank requirement, ensures that none of the original embeddings are eliminated during the fusion. \emph{The second constraint} ensures that each new fused embedding $z'_i$ is predominantly informed by the original $z_i$. Without this requirement, it is hard to ensure accurate prediction of each true label. \emph{The final constraint} ensures that the overall magnitude of each new fused embeddings $z'_i$ matches the magnitude of the original embedding $z_i$. This helps fix an otherwise unconstrained degree of freedom.

% which ensures the diagonal elements of matrix $(I+A)$ are always larger than other elements of the same row, helps 

%Drawing inspiration from~\citep{Yu2023PN}, we apply a linear transformations to the embeddings $Z=\{z_{i},\cdots,z_{\hat(B)}\}$ controlled by design matrix $A$. 

% \begin{align}
% Z &= g_{k}(X) \\
% \tilde{Z} &= (I+A)Z \\
% {\hat{Y}} &= f_{k}(\tilde{Z})
% \end{align}
% where $\hat{Y}$ is the predictions for the images. Subject to the constraints:
% \begin{equation} 
% \begin{split}
% &s.t. \ rank(I+A)=\hat{B} \\
% &\ \quad  \left [ I+A\right ]_{ii} > \left [ I+A \right ]_{ij}, i \ne j \\
% & \ \quad  \left \| \left [ I+A\right ]_{i}   \right \|_{1} = 1 
% \end{split}
% \end{equation}
% Here, the first constraint ensures that $I+A$ is full rank. The full rank requirement ensures that none of the embeddings will be eliminated during the fusion. The second constraint is to make the trained classifier get enough information about a specific image and correctly predict the corresponding label. For example, for an image $x_{1}$ perturbed by another image $x_{2}$, the classifier obtained dominant information from $x_{1}$ so that it can predict the label $y_{1}$. However, if the perturbed image $x_{2}$ is dominant, the classifier can hardly predict the correct label $y_{1}$ and is more likely to predict as $y_{2}$. The third constraint is to maintain the norm of latent representations, keeping the scale of the emebddings unchanged.

\textbf{Concrete design of $A$: Circular Shift.}
Many possible $A$ matrices could satisfy the above desiderata. Here, we adopt a concrete construction of $A$ called \emph{circular shift}~\citep{Yu2023PN}. Under this construction, each $z'_{i+1}$ is perturbed slightly by its immediate previous neighbor $z_{i}$. Because our interdigitated batch layout interleaves labeled and unlabeled samples, this guarantees every labeled embedding is perturbed by an unlabeled embedding.

% This design has been demonstrated to be effective in various \emph{supervised} learning tasks in~\citet{Yu2023PN}. 

% We can formally write out the circular shift fusion strategy as follows.
% Let hyperparameter $\alpha \in (0,0.5)$ define the fusion strength. A circular shift matrix $A$ is formed via $A = \alpha * U - \alpha * I$, where $U_{i,j} = \delta_{i+1,j} $ with $\delta_{i+1,j}$ representing the Kronecker delta indicator \cite{Frankel11Kronecker} and using wrap-around (aka ``circular'') indexing. 
% This yields the overall $Q \times Q$ perturbation matrix:
Formally, let $\alpha \in (0,0.5)$ be the fusion strength. Matrix $A$ is formed via $A = \alpha * U - \alpha * I$, where $U_{i,j} = \delta_{i+1,j} $ with $\delta_{i+1,j}$ representing the Kronecker delta indicator \cite{Frankel11Kronecker} and using wrap-around (aka ``circular'') indexing. 
This yields the overall $Q \times Q$ perturbation matrix:
\begin{align}
 I + A =&
\begin{bmatrix} 
1-\alpha &\alpha  & 0  &  0 & 0 \\
 0 & 1-\alpha & \alpha & 0 & 0 \\
 %0 & 0 & 1-\alpha & \ddots & 0\\
 0 & 0 & \ddots & \ddots & 0 \\
\alpha  & 0 & 0 & 0 &1-\alpha
\end{bmatrix}    
\end{align}
By construction, all three desiderata in Eq.~\eqref{eq:desiderata} are satisfied.

Ultimately this circular shift construction amounts to a simple way to additively perturb each image's embedding with another image's embedding. This simple embedding fusion improves experimental accuracy across many closed-set and open-set SSL tasks in later experiments (Sec.~\ref{Experiments} \& \ref{Experiments_OpenSSL}).

 % \caption{Illustration of Embedding Fusion with Batch Size=2. Embedding Fusion perturbs the feature embeddings by circularly combining them for samples in the minibatch. Showing the flattened embeddings for clear visualization. This process also enables the embeddings of labeled data to incorporate information from the unlabeled set, and vice versa. We demonstrate that this strategy effectively enhances semi-supervised classification performance.}
 
\textbf{Intuition: Why embedding fusion may be desirable.}
The interpolative construction we use is reminiscent of recent strategies for interpolation-based augmentation of embeddings, namely Manifold MixUp~\citep{verma2019manifold}.
Manifold Mixup extends the MixUp idea to the embedding space, linearly interpolating embedding vectors $z_i$ and corresponding label $y_i$.
The authors argue that perturbing the feature embeddings flattens class-specific representations, thus contributing to a smoother decision boundary and better generalization.
Such arguments may also apply to our method. 

% , as discussed further in Sec.~\ref{discussion}.
%More discussion in Sec~\ref{discussion}.
There are several \textbf{key differences between our Embedding Fusion and Manifold Mixup.} First, Manifold MixUp is proposed for supervised learning (not semi-supervised). Procedure-wise, we only perturb the embeddings not the labels; In Manifold MixUp, the interpolation strength is sampled from a Beta distribution, %for each batch,
while we use fixed $\alpha$.
% satisfying eq~\ref{eq:desiderata}. %% MCH commented out, doesn't seem to add much
Most importantly, our interdigitated batch layout deliberately enforces that
each labeled embedding is always blended with an unlabeled
embedding, improving diversity and robustness. 
Careful ablations (see Fig~\ref{fig:CNN_Ablation_FusionStrategy_CIFAR10_40labels}) show this deliberate labeled-unlabeled interaction leads to better classifiers than other layouts with less interaction.

\subsection{Delta Consistency: \emph{Consistent deltas across L \& U}}
\label{RelativeLoss}
% Denoted by 
% $\omega(\cdot)$ and $\Omega(\cdot)$
%  respectively as a predefined set of weak and strong augmentation operations, $\omega(\cdot)_{m}$ and $\Omega(\cdot)_{m}$ as a random sample from $\omega(\cdot)$ and $\Omega(\cdot)$.
% For a labeled sample $x_{l}^{(i)} \in \mathcal{L}$, we generate a weakly augmented sample $x_{l}^{(i)w}$ and a strongly augmented sample $x_{l}^{(i)s}$ using $\omega(\cdot)_{m}$ and $\Omega(\cdot)_{m}$. For each unlabeled sample $x_{u}^{(j)} \in \mathcal{U}, j \in \{1,\ldots,\mu\}$, where $\mu$ is the ratio of unlabeled data batch size to labeled data batch size as in~\citep{sohn2020fixmatch}, we apply the \textit{same} weak and strong augmentation $\omega(\cdot)_{m}$ and $\Omega(\cdot)_{m}$ to obtain $x_{u}^{(j)w}$, and $x_{u}^{(j)s}$, $j\in\{1,\ldots,\mu\}$. 
%The difference in augmentation $\omega(\cdot)_{m}$ and $\Omega(\cdot)_{m}$ results in the difference in predictions for the same data.

We now introduce the second key contribution of our  InterLUDE: a new loss we call \emph{delta consistency
loss}. Delta consistency
loss makes use of the widely-used consistency regularization principle~\citep{tarvainen2017mean,sohn2020fixmatch}. However, as noted in prior work~\citep{huang2023flatmatch}, most consistency-regularization losses only encourage ``instance-wise'' consistency, that is, 
consistency for \emph{each individual instance} under different transformations. In contrast, our delta consistency
loss is designed to make deltas (changes) in class-prediction behavior consistent \emph{across} labeled (L) and unlabeled (U) instances.

%To understand relative loss, 
Recall that our overall approach (see Alg.~\ref{alg:iMatch}) applies the same weak and strong augmentation to adjacent sets of $1+\mu$ images (1 labeled image and $\mu$ unlabeled images). The key idea is that, given a specific pair of weak and strong augmentations $\omega$ and $\sigma$, the \textbf{change in predicted probabilities due to swapping weak with strong augmentation should be similar} across labeled and unlabeled cases. Intuitively, if a specific augmentation swap causes a labeled image to look more like ``dog'' and less like ``cat'', it ought to produce a similar change for unlabeled images \emph{on average}.

Define index $i \in \{1, \ldots B\}$ to uniquely identify a distinct labeled image in the current batch.
Let $\omega_i, \sigma_i$ be the specific weak-strong pair of augmentations to be swapped for index $i$. 
Denote $p_i^w$ (respectively $p_i^s$) as the class probabilities produced by classifier $h$ given the weak embedding $z^{\prime,w}_i$ (strong embedding $z^{\prime,s}_i$). 
Let index $m \in \{1, 2, \ldots \mu\}$ denote an offset from $i$, so subscript $i,m$ identifies an unlabeled example that is adjacent to $i$ in our interdigitated layout.
Let vector $q_{i,m}^w$ (respectively $q_{i,m}^s$) denote the  class probabilities produced by classifier $h$ given the weak embedding $z_{i,m}^{\prime,w}$ (strong embedding $z_{i,m}^{\prime,s}$) of that unlabeled image.

% We now define our relative loss. Recall that our approach (see Alg.~\ref{alg:iMatch}) produces a weak and strong augmentation for each image (labeled and unlabeled) in the batch.
% Furthermore, by construction the same weak augmentation $\omega$ and same strong augmentation $\sigma$ is applied to adjacent sets of $1+\mu$ rows in the embedding array $Z$.
% The key idea of relative loss is that, when using the same weak and strong augmentation (realization $\omega$ and $\sigma$), the \textbf{change in predicted probabilities due to swapping weak with strong augmentation} should be similar across images. Intuitively, if a specific augmentation swap causes a labeled image to look more like ``dog'' and less like ``cat'', it ought to produce similar change for the unlabeled images on average.

% Define index $i \in \{1, \ldots B\}$ to unique identify a distinct labeled image within the current batch.
% Denote $p_i^w$ (respectively $p_i^s$) as the $C$-dimensional probability vector produced by applying the classification head $h$ to the weak embedding $z_i$ (strong embedding $z^s_i$) of example $i$. 
% Given an index $i$, let index $m \in \{1, 2, \ldots \mu\}$ denote an offset from $i$, describing an unlabeled example that is adjacent to $i$ in our interdigitated layout.
% Let vector $q_{i+m}^w$ (respectively $q_{i+m}^s$) denote the $C$-dimensional probability vector produced by classifier $h$ given the weak embedding $z_{i+m}^w$ (strong embedding $z_{i+m}^s$). 

Next, define vectors $\Delta^L_i,\Delta^U_i$ that represent the differences of predicted probabilities for the labeled and unlabeled case:
% \begin{align}
% \Delta^L_i &= p_{i}^w - p_{i}^s, & \Delta^U_i &= \frac{1}{\mu} \sum_{m=1}^{\mu} (q_{i+m}^w - q_{i+m}^s). \label{eq:diff_labeled}
% \end{align}
\begin{align}
\Delta^L_i &= p_{i}^w - p_{i}^s, \label{eq:diff_labeled}
\quad 
\Delta^U_i = \frac{1}{\mu} \sum_{m=1}^{\mu} (q_{i,m}^w - q_{i,m}^s).
\end{align}
The goal of delta consistency
loss is to encourage $\Delta^U_i$, the average change in predictions across unlabeled instances associated with $i$, to mimic the change on the corresponding labeled instance $\Delta^L_i$.
Concretely, we minimize the average squared Euclidean distances between vectors $\Delta^L_i$ and $\Delta^U_i$:
\begin{align}
    \mathcal{L}^{\text{DC}}= \frac{1}{B}\sum_{i=1}^{B} \left\| \Delta_i^L - \Delta_i^U \right\|_2^2 \label{eq:loss_rel}
\end{align}
Many possible distance functions could be tried; we picked this distance because it is simple, symmetric, and bounded. The bounded property may help prevent fluctuation in training dynamics~\citep{berthelot2019mixmatch}.

\begin{algorithm}[h]
\caption{InterLUDE and InterLUDE+}
\label{alg:iMatch}
\textbf{Input}: Labeled set $\mathcal{D}^L$, Unlabeled set $\mathcal{D}^U$ 
\\
\textbf{Output}: Trained weights $\theta$ for classifier $f$
% \textbf{Hyperparam.}: 
%     $\alpha \in (0,.5)$: fusion strength
% \\
% ~~~~$\mu$ : multiplier for size of unlabeled
% \\
% ~~~~$\epsilon$: step size for SGD
% \\
% ~~~~$\Omega, \Sigma$: possible weak and strong augmentations
\\
\textbf{Procedure} 
\linespread{1.1}\selectfont % improve spacing between lines
\begin{algorithmic}[1] %[1] enables line numbers
\FOR{iter $t \in 1, 2, \ldots T$}

\STATE $\{ x_i, y_i \}_{i=1}^B \gets \textsc{DrawBatch}(\mathcal{D}^L, B)$
\STATE $\{ \bar{x}_j \}_{j=1}^{\mu B} \gets \textsc{DrawBatch}(\mathcal{D}^U, \mu B)$
%\STATE $\{(x_{l}^{(i)},y^{(i)})\}^B, \{(x_{u}^{(j)})\}^{\mu B} \gets {\small \textsc{Sample Minibatch}}(\mathcal{D}^L, \mathcal{D}^U)$

%\STATE $Z \gets \textsc{Zeros}( ((\mu+1)B, D))$
\FOR{example $i \in 1, 2, \ldots B$}
\STATE $\omega_{i}, \sigma_{i} \gets {\small \textsc{DrawAugParams}}(\Omega, \Sigma)$
\STATE $x^w_i, x^s_i, \{\bar{x}^w_j\}_{j=1}^{\mu}, \{\bar{x}^s_i\}_{j=1}^{\mu} \gets \textsc{GetAug}(\omega_i, \sigma_i)$

\ENDFOR
% \STATE $\{x^w_i\}_{i=1}^B, \{x^s_i\}_{i=1}^B, \{\bar{x}^w_j\}_{j=1}^{\mu B}, \{\bar{x}^s_i\}_{j=1}^{\mu B} \gets \textsc{Aug}(\{\omega_i\}_{i=1}^B, \{\sigma_i\}_{i=1}^B)$

% \STATE $\{x^w_i\}_{i=1}^B \gets \textsc{WeakAug}( \{x_i\}_{i=1}^B, \{\omega_i\}_{i=1}^B )$
% \STATE $\{x^s_i\}_{i=1}^B \gets \textsc{StrongAug}( \{x_i\}_{i=1}^B, \{\sigma_i\}_{i=1}^B )$
% \STATE $\{\bar{x}^w_j\}_{j=1}^{\mu B} \gets \textsc{WeakAug}( \{\bar{x}_j\}_{j=1}^{\mu B}, \textsc{Tile}(\{\omega_i\}, \mu) ) )$
% \STATE $\{\bar{x}^s_i\}_{j=1}^{\mu B} \gets \textsc{StrongAug}( \{\bar{x}_j\}_{j=1}^{\mu B}, \textsc{Tile}(\{\sigma_i\}, \mu) )$

%\STATE $x^w_i \gets \textsc{WeakAug}( x_i, \omega_i )$
%\STATE $x^s_i \gets \textsc{StrongAug}( x_i, \Omega_i)$
%\FOR{example $m \in 1, 2, \ldots \mu$}
%    \STATE $\bar{x}^w_{i+m} \gets \textsc{WeakAug}( x_i, \omega_i )$
%    \STATE $\bar{x}^s_{i+m} \gets \textsc{StrongAug}( x_i, \Omega_i)$
%\ENDFOR
%\STATE $\{(x_{l}^{(i)w},x_{l}^{(i)s})\}, \{(x_{u}^{(j)w}, (x_{u}^{(j)s})\}^{\mu} \gets {\small \textsc{Apply Augmentation}}(\omega_{m}, \Omega_{m})$
%\ENDFOR

\STATE $X \gets \textsc{InterDigitate}( \{x^w_i, x^s_i\}_{i=1}^B, \{ \bar{x}^w_j, \bar{x}^s_j\}_{j=1}^{\mu B} )$

\STATE $Z \gets g(X; \theta)$ \text{~~~// calc embeddings}
\STATE $Z' \gets \textsc{CircShiftFusion}(Z, \alpha)$
\STATE $\{p_{i}^w, p_{i}^s\}_{i=1}^B, \{q_{j}^w, q_{j}^s\}_{j=1}^{\mu B} \gets h( Z'; \theta)$ %\textsc{Predict}(Z')$
%\STATE $ \{p_{i}^w, p_{j}^s\}^B, \{q_{j}^w, q_{j}^s\}^{\mu B}
%\gets \textsc{Prediction with Embedding Fusiosn}$

\STATE $ \mathcal{L}^{\text{DC}}
\gets \textsc{Eq.}~\eqref{eq:loss_rel}$
\text{~~ // delta-consistency loss}
\STATE $ \mathcal{L}^{\text{L}}
\gets \textsc{Eq.}~\eqref{eq:supervised_loss}$
\text{~~~ // supervised cross-ent. loss}
\STATE $ \mathcal{L}^{\text{U}}
\gets \textsc{Eq.}~\eqref{eq:instance_wise_unlabeled_loss}$ \text{~~~ // instance-wise unlabeled loss}

\IF{InterLUDE}
\STATE $\mathcal{L} \gets \mathcal{L}^{\text{L}} + \lambda_{\text{u}} \mathcal{L}^{\text{U}} + \lambda_{\text{DC}} \mathcal{L}^{\text{DC}}$
\ELSIF{InterLUDE+}
\STATE $ \mathcal{L}
\gets \textsc{Eq.~\eqref{eq:imatchplus_total_loss}}$
\ENDIF

% \STATE $ \mathcal{L}_{\text{u}}
% \gets \textsc{Eq.}~\eqref{eq:Lu}$ \text{~~~ // unsupervised loss}

% \STATE $\mathcal{L} \gets \mathcal{L}_{\text{sup}} + \lambda_{\text{rel}} \mathcal{L}_{\text{rel}} + \lambda_{\text{u}} \mathcal{L}_{\text{u}} $
% \IF{iMatch+}
% \STATE $ \mathcal{L}
% \gets \textsc{eq~\ref{eq:imatchplus_total_loss}}$
% \ENDIF

\STATE $\theta = \theta - \epsilon \nabla_{\theta} \mathcal{L}$ \text{~~~ // update weights via SGD} 

\ENDFOR
\STATE \textbf{return} $\theta$
\end{algorithmic}
% \textbf{Hyperparameters: Details in Experiment Sections}
% \textbf{Hyperparameters} (\emph{Values marked $\dagger$ tuned for all baselines as in App.~\ref{app:hyperparameter_list}. No tuning for Fix-A-Step in any experiment.})
% \begin{itemize}[align=left,style=nextline,leftmargin=*,labelsep=3\parindent,font=\normalfont,topsep=0pt,itemsep=-1ex,partopsep=1ex,parsep=1ex]
% 	\item ~Temperature $\tau{=}0.5$ for {\small \textsc{Aug+PseudoLabel}} (Alg.~\ref{alg:aug_softlabel})
% 	\item ~Beta dist. shape $\alpha{=}0.5$ for {\small \textsc{MixMatchAug}} (Alg.~\ref{alg:mixmatch})
% 	\item ~Step size $\epsilon$$^\dagger$, Initial weights $w$, Max iterations $I$
% 	\item ~Unlabeled-loss weight per iter $\lambda_1, \ldots \lambda_I$$^\dagger$
% \end{itemize}
\end{algorithm}
 
\subsection{InterLUDE overall training objective}

Ultimately, our proposed InterLUDE algorithm combines the classic two-term SSL objective from Eq.~\eqref{eq:standard_ssl_objective} with our new delta consistency loss.
Our final loss function is: 
\begin{align}
\label{eq:imatch_total_loss}
\mathcal{L} &= \mathcal{L}^L + \lambda_u \mathcal{L}^U + \lambda_{\text{DC}} \mathcal{L}^{\text{DC}}
\end{align}
where $\lambda_{u} > 0$ and $\lambda_{DC}>0$ control the relative weight of different terms. We use the following common choices for labeled loss $\mathcal{L}^L$ and unlabeled loss $\mathcal{L}^U$
\begin{align}
\mathcal{L}^L &= -\frac{1}{B} \sum_{i=1}^{B} y_{i} \log p_{i}^w \label{eq:supervised_loss}
\\
 \mathcal{L}^U &= \frac{1}{\mu B} \sum_{j=1}^{\mu B} \mathbf{1} (\max (q_{j}^w) > \tau) H(q_{j}^w, q_{j}^s)\label{eq:instance_wise_unlabeled_loss}
\end{align}
Here, $\mathcal{L}^L$ is a standard cross-entropy loss based on labeled samples only; $\mathcal{L}^U$ is the instance-wise consistency loss exemplified by FixMatch ~\citep{sohn2020fixmatch} that has become a useful part of the final objectives of many methods~\citep{huang2023flatmatch,wang2022freematch,chen2023softmatch}. We fix confidence threshold $\tau$ to 0.95, following~\citeauthor{sohn2020fixmatch}.

% , is defined as
% \begin{align}
% \label{eq:Lu}
%  \mathcal{L}_u = \frac{1}{\mu B} \sum_{j=1}^{\mu B} \mathbf{1} (\max (q_{j}^w) > \tau) H(q_{j}^w, q_{j}^s)
%  \end{align}
% where $\tau$ is a confidence threshold, which we default to 0.95 following~\citeauthor{sohn2020fixmatch}. This widely-used unlabeled loss is a key part of the final objectives of many previous methods~\citep{huang2023flatmatch,wang2022freematch,zheng2022simmatch,zheng2023simmatchv2}.

% We use standard cross-entropy loss only on the weakly augmented labeled samples. Denote $p(y|x_{i}^w)$ as $p_{i}^w$
% Our final loss function can be written as: 
% \begin{align}
% \mathcal{L}_s &= -\frac{1}{B} \sum_{i=1}^{B} y_{i} \log p_{i}^w \\
% \mathcal{L} &= \mathcal{L}_{s} + \lambda_{u}\mathcal{L}_u + \lambda_{rel} \mathcal{L}_{rel}
% \end{align}

% where $\lambda_{u}$ and $\lambda_{rel}$ are balancing factors that control the weights of the different losses.

\subsection{InterLUDE+}
Recently, the Self-Adaptive Threshold (SAT) and Self-Adaptive Fairness (SAF) ideas were introduced by~\citet{wang2022freematch}, with possible applicability to other SSL algorithms. For instance, FlatMatch~\cite{huang2023flatmatch} has utilized these techniques. 
For fair comparison to such past work, we integrate these techniques into InterLUDE and name the enhanced version InterLUDE+. 
% Note that iMatch achieves superior performance to FlatMatch in many scenarios and remains competitive in others, even without the incorporation of SAT and SAF.
We provide a brief overview of SAT and SAF here; more details in App.~\ref{app_iMatch+_details}.

SAT is a technique for adjusting the pseudo-labels threshold over iterations based on both dataset-specific and class-specific criteria. %Therefore, the loss term for the unlabeled data assigned with pseudo-labels will be impacted by SAT.
%at different time step.
%To incorporate SAT, a dynamic threshold $\tau_t(c)$ is used in eq.~\ref{eq:Lu} (instead of fixed $\tau$). $c$ and $t$ denotes the class and current time step. 
SAF is a regularization technique that encourages diverse predictions on unlabeled data via an additional loss term $\mathcal{L}^{\text{SAF}}$.
Thus, the complete loss function of InterLUDE+ (including SAT and SAF) is: 
\begin{align}
\mathcal{L} &= \mathcal{L}^{L} + \lambda_{u}\mathcal{L}^{U} + \lambda_{\text{DC}} \mathcal{L}^{\text{DC}} + \lambda_{\text{SAF}} \mathcal{L}^{\text{SAF}} 
\label{eq:imatchplus_total_loss}
\end{align}
with a self-adaptive threshold replacing the fixed $\tau$ in $L^{U}$.

\section{Experiments on Classic SSL Benchmarks}
\label{Experiments}
We evaluate InterLUDE on common closed-set SSL benchmarks with both Convolutional Neural Network (CNN) and Vision Transformer (ViT, \citet{dosovitskiy2020image}). Our comparisons encompass a \textbf{comprehensive list of 21 SSL algorithms}, including recent state-of-the-art methods. 
%Our extensive experiments demonstrate that iMatch significantly advances current SSL performance.

% In this section, we conduct experiments on common SSL benchmarks adopted by most SSL papers. Beside the commonly used Convolutional Neural Network (CNN) architectures, we also include the more advanced Vision Transformers (ViT) architecture~\citep{dosovitskiy2020image}. We compare to a comprehensive list of \textbf{21 SSL algorithms} including {the \textbf{most recent SOTAs}. Through extensive experiments, we show that iMatch substaintially advance current SSL. Our open source code\footnote{\todo{code will be available upon acceptance}} uses PyTorch \citep{paszke2019pytorch} and allows reproducing each experiments in the paper. 

\textbf{Datasets.}
 We use CIFAR-10~\citep{krizhevsky2009learning}, CIFAR-100 ~\citep{krizhevsky2009learning} and STL-10~\citep{coates2011analysis}. Detailed descriptions can be found in App.~\ref{app_dataset_details}.
We skip SVHN~\citep{netzer2011reading} as recent SSL advances have pushed performance to near saturation (e.g., ${<}2\%$ error rate with only 4 labels per class~\citep{wang2022freematch}). 

%hz: can be moved to appendix
% \begin{itemize}
%     \item CIFAR-10  consists of 60K 32x32 color images in 10 classes, with 6k images per class. The training and test set contains 5K and 1k samples, respectively. 
    
%     \item CIFAR-100  contains 100 classes with 600 images each. There are 500 training images and 100 testing images per class. 
    
%     \item STL-10 consists of 5K labeled images of size 96x96, 100K unlabeled images, and 8,000 test images. 
% \end{itemize}

\textbf{CNN experiment setting.} Using CNNs as backbones, we conduct experiments across various labeled set sizes: 40, 250, 4000 for CIFAR-10; 400, 2500, 10000 for CIFAR-100 and 250, 1000 for STL-10.  
Following standard protocol~\citep{wang2022freematch,zheng2023simmatchv2}, we use Wide ResNet-28-2/28-8~\citep{zagoruyko2016wide} for CIFAR-10/100 and ResNet-37-2~\citep{he2016deep} for STL10. We use SGD optimizer (Nesterov momentum 0.9) and a cosine learning rate schedule~\citep{loshchilov2016sgdr} $\eta = \eta_0 \cos\left(\frac{7\pi k}{16K}\right)$, where $\eta_{0}=0.03$ is the initial learning rate, $K=2^{20}$ is the total training steps and $k$ is the current training step. 
%Initial learning rate is set to 0.03 and the total training s set to $2^{20}$. 
Inference is conducted using the exponential moving average of the model with a momentum of 0.999. We set the labeled batch size to 64 and the unlabeled batch size to 448 (i.e., $\mu=7$). For hyperparameters unique to InterLUDE, we set $\lambda_{DC}$ to 1.0 and fusion strength $\alpha$ to 0.1.

\textbf{ViT experiment setting.}
Using ViTs as backbones, we conduct experiments across labeled set sizes of 10, 40, 250 for CIFAR-10; 200, 400, 2500 for CIFAR-100; and 10, 40, and 100 for STL-10. Following prior works~\citep{Wang2022USB, li2023instant}, we use ViT-small for CIFAR-10/100 and ViT-base for STL-10. We use AdamW optimizer with learning rate $5e^{-4}$ for CIFAR-10/100 and $1e^{-4}$ for STL-10. The total training steps are set to 204,800. The batch size is set to 8 with $\mu=7$. For hyperparameters unique to InterLUDE, we set $\lambda_{DC}$ to 0.1 and fusion strength $\alpha$ to 0.1. 

\textbf{Results.} Results are shown in 
Table~\ref{wideresnet-table} for CNNs and Table~\ref{vit-compa} for pretrained ViTs. It is worth noting that previous observers suggest it is hard for an algorithm to achieve top performance across all settings~\citep{zheng2023simmatchv2}. 

On CNN backbones (Table~\ref{wideresnet-table}), 
our methods achieve the best results in many scenarios and remain competitive in others. Notably, on CIFAR-10 with 40 labels, InterLUDE records a low error rate of 4.51\%, which InterLUDE+ further pushes to 4.46\%. InterLUDE and InterLUDE+ thus do not only outperform SSL alternatives but are the \textbf{only methods to surpass the fully-supervised result}, despite using only 40 labeled images.
Moreover, InterLUDE is consistently more stable than competitors like FlatMatch (see much smaller standard deviations in Tab.~\ref{wideresnet-table} and later results in Tab.~\ref{tab:Heart2Heart}).

On ViT backbones (Table~\ref{vit-compa}), InterLUDE+ delivers the \textbf{best performance across almost all scenarios}, with InterLUDE close behind. On STL-10 with only 40 labels, InterLUDE (InterLUDE+) achieves an impressive low \textbf{error rate of 3.14\% (4.59\%), far better than the next best method of 14.91\% error rate} and remarkably better than competitors even when they have  2.5x more labeled images (last column).
On STL-10 with 100 labels, InterLUDE achieves over 3x lower error rate compared to runner-ups.  
% iMatch improves further with 100 labeled images on STL-10, achieving over \todo{3x lower error rate} compared to runner-ups.  
%, marking a significant milestone in SSL. 
%% MCH commented this out, we don't want to brag too much
%Even with more 100 labels, iMatch (iMatch+) continue

Comparing across Tables 1-2, ViT backbones exhibit superior performance, in line with observations in~\citet{Wang2022USB}. InterLUDE and InterLUDE+ perform similarly on CNNs, but with ViTs InterLUDE+ appears better on CIFAR.%-10/100.

% Comparing ViT backbones and CNN backbones, ViT backbone demonstrated superior performance, align with the observation in USB benchmark~\citep{Wang2022USB}. Interestingly, iMatch and iMatch+ show similar performances on CNNs, while iMatch+ appears more effective on ViTs. More discussion in App~\ref{app_additional_Discussion}.

% We hypothesize this is due to the behavior of the adopted thresholding strategy SAT~\citep{wang2022freematch} in iMatch+. More discussion in App~\ref{app_additional_Discussion}.

% , which are copied from the USB benchmarks~\citep{Wang2022USB}
%SimMatch version
\begin{table*}[h]
\caption{Error rate (\%) with CNNs. Following~\citep{zheng2023simmatchv2}, error rate and standard deviation are reported based on three runs. All experiments follow the same settings. Results of other methods are directly copied from SimMatchV2~\citep{zheng2023simmatchv2} and the original papers (``--'' means the result is not available). The best results are highlighted in bold and the second-best underlined.}
\label{wideresnet-table}
\resizebox{\textwidth}{!}{\begin{tabular}{ccccccccc}
\hline
\textbf{Dataset}                     & \multicolumn{3}{c}{\textbf{CIFAR10}}                                                                         & \multicolumn{3}{c}{\textbf{CIFAR100}}                                                                        & \multicolumn{2}{c}{\textbf{STL-10}}                                                     \\ \hline
\multicolumn{1}{c|}{\#Label}         & 40                          & 250                         & \multicolumn{1}{c|}{4000}                        & 400                         & 2500                        & \multicolumn{1}{c|}{10000}                                              & 250                         & 1000                        \\ \hline
\multicolumn{1}{c|}{Fully-Supervised}       & & 4.57$\pm$0.06 &  & & 18.96$\pm$0.06 &  &  \multicolumn{2}{c}{--}                     \\ \hline

\multicolumn{1}{c|}{Supervised}       & 77.18$\pm$1.32 & 56.24$\pm$3.41 & \multicolumn{1}{c|}{16.10$\pm$0.32} & 89.60$\pm$0.43 & 58.33$\pm$1.41 & \multicolumn{1}{c|}{36.83$\pm$0.21} &  55.07$\pm$1.83 & 35.42$\pm$0.48                     \\ \hline

\multicolumn{1}{c|}{Pseudo-Labeling~\citep{lee2013pseudo}} & 75.95$\pm$1.86 & 51.12$\pm$2.91 & \multicolumn{1}{c|}{15.32$\pm$0.35} & 88.18$\pm$0.89 & 55.37$\pm$0.48 & \multicolumn{1}{c|}{36.58$\pm$0.12} &  51.90$\pm$1.87 & 30.77$\pm$0.04 \\
\multicolumn{1}{c|}{II-Model~\citep{laine2016temporal}}        & 76.35 $\pm$ 1.69 & 48.73$\pm$1.07 & \multicolumn{1}{c|}{13.63$\pm$0.07} & 87.67$\pm$0.79 & 56.40$\pm$0.69 & \multicolumn{1}{c|}{36.73$\pm$0.05} &  52.20$\pm$2.11 & 31.34$\pm$0.64 \\
\multicolumn{1}{c|}{Mean Teacher~\citep{tarvainen2017mean}}    & 72.42$\pm$2.10 & 37.56$\pm$4.90 & \multicolumn{1}{c|}{8.29$\pm$0.10} & 79.96$\pm$0.53 & 44.37$\pm$0.60 & \multicolumn{1}{c|}{31.39$\pm$0.11} &  49.30$\pm$2.09 & 27.92$\pm$1.65 \\
\multicolumn{1}{c|}{VAT~\citep{miyato2018virtual}}             & 78.58$\pm$2.78 & 28.87$\pm$3.62 & \multicolumn{1}{c|}{10.90$\pm$0.16} & 83.60$\pm$4.21 & 46.20$\pm$0.80 & \multicolumn{1}{c|}{32.14$\pm$0.31} &  57.78$\pm$1.47 & 40.98$\pm$0.96 \\
\multicolumn{1}{c|}{MixMatch~\citep{berthelot2019mixmatch}}        & 35.18$\pm$3.87 & 13.00$\pm$0.80 & \multicolumn{1}{c|}{6.55$\pm$0.05} & 64.91$\pm$3.34 & 39.29$\pm$0.13 & \multicolumn{1}{c|}{27.74$\pm$0.27} &  32.05$\pm$1.16 & 20.17$\pm$0.67 \\
\multicolumn{1}{c|}{ReMixMatch~\citep{berthelot2019remixmatch}}      & 8.13$\pm$0.58  & 6.34$\pm$0.22 & \multicolumn{1}{c|}{4.65$\pm$0.09} & 41.60$\pm$1.48 & 25.72$\pm$0.07 & \multicolumn{1}{c|}{20.04$\pm$0.13} &  11.14$\pm$0.52 & 6.44$\pm$0.15 \\
\multicolumn{1}{c|}{FeatMatch~\citep{kuo2020featmatch}}      & --  & 7.50$\pm$0.64 & \multicolumn{1}{c|}{4.91$\pm$0.18} & -- & -- & \multicolumn{1}{c|}{--} &  -- & -- \\
\multicolumn{1}{c|}{UDA~\citep{xie2020unsupervised}}             & 10.01$\pm$3.34 & 5.23$\pm$0.08 & \multicolumn{1}{c|}{4.36$\pm$0.09} & 45.48$\pm$0.37 & 27.51$\pm$0.28 & \multicolumn{1}{c|}{23.12$\pm$0.45} &  10.11$\pm$1.15 & 6.23$\pm$0.28 \\
\multicolumn{1}{c|}{FixMatch~\citep{sohn2020fixmatch}}        & 12.66$\pm$4.49 & 4.95$\pm$0.10 & \multicolumn{1}{c|}{4.26$\pm$0.01} & 45.38$\pm$2.07 & 27.71$\pm$0.42 & \multicolumn{1}{c|}{22.06$\pm$0.10} & 8.64$\pm$0.84 & 5.82$\pm$0.06 \\
\multicolumn{1}{c|}{Dash~\citep{xu2021dash}}            & 9.29$\pm$3.28  & 5.16$\pm$0.28 & \multicolumn{1}{c|}{4.36$\pm$0.10} & 47.49$\pm$1.05 & 27.47$\pm$0.38 & \multicolumn{1}{c|}{21.89$\pm$0.16} & 10.50$\pm$1.37 & 6.30$\pm$0.49 \\
\multicolumn{1}{c|}{MPL~\citep{pham2021meta}}            & 6.62$\pm$0.91  & 5.76$\pm$0.24 & \multicolumn{1}{c|}{4.55$\pm$0.04} & 46.26$\pm$1.84 & 27.71$\pm$0.19 & \multicolumn{1}{c|}{21.74$\pm$0.09} & --  & 6.66$\pm$0.00 \\
\multicolumn{1}{c|}{CoMatch~\citep{li2021comatch}}         & 6.51$\pm$1.18  & 5.35$\pm$0.14 & \multicolumn{1}{c|}{4.27$\pm$0.12} & 53.41$\pm$2.36 & 29.78$\pm$0.11 & \multicolumn{1}{c|}{22.11$\pm$0.22} &  7.63$\pm$0.94 & 5.71$\pm$0.08 \\
\multicolumn{1}{c|}{FlexMatch~\citep{zhang2021flexmatch}}       & 5.29$\pm$0.29  &  4.97$\pm$0.07 & \multicolumn{1}{c|}{4.24$\pm$0.06} & 40.73$\pm$1.44 & 26.17$\pm$0.18 & \multicolumn{1}{c|}{21.75$\pm$0.15} & 9.85$\pm$1.35 & 6.08$\pm$0.34 \\
\multicolumn{1}{c|}{AdaMatch~\cite{berthelot2021adamatch}}        & 5.09$\pm$0.21  & 5.13$\pm$0.05 & \multicolumn{1}{c|}{4.36$\pm$0.05} & 37.08$\pm$1.35 & 26.66$\pm$0.33 & \multicolumn{1}{c|}{21.99$\pm$0.15} & 8.59$\pm$0.43 & 6.01$\pm$0.02 \\
\multicolumn{1}{c|}{SimMatch~\citep{zheng2022simmatch}}        & 5.38$\pm$0.01  & 5.36$\pm$0.08 & \multicolumn{1}{c|}{4.41$\pm$0.07} & 39.32$\pm$0.72 & 26.21$\pm$0.37 & \multicolumn{1}{c|}{21.50$\pm$0.11} & 8.27$\pm$0.40 & 5.74$\pm$0.31 \\
\multicolumn{1}{c|}{FreeMatch~\citep{wang2022freematch}}      & 4.90$\pm$0.04  & 4.88$\pm$0.18 & \multicolumn{1}{c|}{4.10$\pm$0.02} & 37.98$\pm$0.42 & 26.47$\pm$0.20 & \multicolumn{1}{c|}{21.68$\pm$0.03} & -- & 5.63$\pm$0.15 \\
\multicolumn{1}{c|}{SoftMatch~\citep{chen2023softmatch}}      & 4.91$\pm$0.12  & 4.82$\pm$0.09 & \multicolumn{1}{c|}{4.04$\pm$0.02} & 37.10$\pm$0.77 & 26.66$\pm$0.25 & \multicolumn{1}{c|}{22.03$\pm$0.03} &  -- & 5.73$\pm$0.24 \\
\multicolumn{1}{c|}{SimMatchV2~\citep{zheng2023simmatchv2}}      & 4.90$\pm$0.16  & 5.04$\pm$0.09 & \multicolumn{1}{c|}{4.33$\pm$0.16} & \underline{36.68$\pm$0.8}6 & 26.66$\pm$0.38 & \multicolumn{1}{c|}{21.37$\pm$0.20} & 7.54$\pm$0.81 & 5.65$\pm$0.26 \\
\multicolumn{1}{c|}{FixMatch (w/SAA)~\citep{gui2023enhancing}}        & 5.24$\pm$0.99  & 4.79$\pm$0.07 & \multicolumn{1}{c|}{3.91$\pm$0.07} & 45.71$\pm$0.73 & 26.82$\pm$0.21 & \multicolumn{1}{c|}{21.29$\pm$0.20} & 
-- & -- \\
\multicolumn{1}{c|}{InstanT~\citep{li2023instant}}        & 5.17$\pm$0.10  & 5.28$\pm$0.02 & \multicolumn{1}{c|}{4.43$\pm$0.01} & 46.06$\pm$1.80 & 32.91$\pm$0.00 & \multicolumn{1}{c|}{27.70$\pm$0.40} & 
-- & -- \\
\multicolumn{1}{c|}{FlatMatch~\citep{huang2023flatmatch}}      & 5.58$\pm$2.36  & \textbf{4.22$\pm$1.14} & \multicolumn{1}{c|}{\underline{3.61$\pm$0.49}} & 38.76$\pm$1.62 & 25.38$\pm$0.85 & \multicolumn{1}{c|}{\textbf{19.01$\pm$0.43}} & 
-- & \textbf{4.82$\pm$1.21} \\
\multicolumn{1}{c|}{FlatMatch-e~\citep{huang2023flatmatch}}      & 5.63$\pm$1.87  & 4.53$\pm$1.85 & \multicolumn{1}{c|}{\textbf{3.57$\pm$0.50}} & 38.98$\pm$1.53 & 25.62$\pm$0.88 & \multicolumn{1}{c|}{\underline{19.78$\pm$0.89}} & 
-- & 5.03$\pm$1.06 \\
\hline
\multicolumn{1}{c|}{InterLUDE (ours)}          &\underline{4.51$\pm0.01$}                             &                             4.63$\pm$0.11 &\multicolumn{1}{c|}{3.96$\pm$0.07}                            &                             \textbf{35.32$\pm$1.06} &  \textbf{25.20$\pm$0.22}                            & \multicolumn{1}{c|}{20.77$\pm$0.19}                            &                            \underline{7.05$\pm$0.12}                          & 5.01$\pm$0.04\\     \multicolumn{1}{c|}{InterLUDE+ (ours)}          &\textbf{4.46$\pm$0.11}                             &                             \underline{4.46$\pm0.09$} &\multicolumn{1}{c|}{3.88$\pm0.05$}                            &36.99$\pm$0.62                              &\underline{25.27$\pm$0.17}                              & \multicolumn{1}{c|}{20.49 $\pm$0.15}                            &\textbf{6.99 $\pm$0.42}                                                      & \underline{4.92$\pm$0.05}\\   \hline                          
\end{tabular}}
\end{table*}

\begin{table*}[ht]
\caption{Error rate (\%) with ViT backbone. The error rate and 95\% confidence interval are reported based on three random seeds~\cite{li2023instant}. Other results directly copied from~\citet{li2023instant}. The best results are highlighted in bold and the second-best underlined.
}
\resizebox{\textwidth}{!}{\begin{tabular}{c|lll|ccc|ccc}
\hline
Dataset                             & \multicolumn{3}{c|}{CIFAR10}                                               & \multicolumn{3}{c|}{CIFAR100}                                       & \multicolumn{3}{c}{STL10}                                          \\ \hline
\#Label                             & \multicolumn{1}{c}{10} & \multicolumn{1}{c}{40} & \multicolumn{1}{c|}{250} & 200                  & 400                  & 2500                  & 10                   & 40                   & 100                  \\ \hline
PL   \cite{lee2013pseudo}                                & 62.35$\pm$3.1          & 11.79$\pm$5.3          & 4.58$\pm$0.4            &36.66$\pm$2.0       & 26.87$\pm$0.9        & 15.72$\pm$0.1         & 69.26$\pm$6.7        & 42.84$\pm$4.2        & 26.56$\pm$1.5        \\
MT    \cite{tarvainen2017mean}                                & 35.43$\pm$4.9          & 12.85$\pm$2.5          & 4.75$\pm$0.5            & 40.5$\pm$ 0.8      &30.58$\pm$0.9        & 17.09$\pm$0.4         &57.28$\pm$7.8        & 33.20$\pm$3.4        & 22.29$\pm$1.8        \\
MixMatch   \cite{berthelot2019mixmatch}                         & 34.96$\pm$2.6          & 2.84$\pm$0.9          & 2.05$\pm$0.1            &39.64$\pm$ 1.3       & 27.74$\pm$0.1        & 16.16$\pm$0.3         & 89.32$\pm$1.1        & 72.42$\pm$16.2       & 38.15$\pm$11.3       \\
VAT    \cite{miyato2018virtual}                               & 39.93$\pm$6.3          & 6.67$\pm$6.6          & 2.33$\pm$0.2            & 34.11$\pm$1.8        & 24.67$\pm$0.4        & 16.58$\pm$0.4         & 79.43$\pm$4.4        & 34.82$\pm$7.0        & 19.06$\pm$1.0        \\
UDA    \cite{xie2020unsupervised}                                & 21.24$\pm$3.6          & 2.08$\pm$0.2          & 2.04$\pm$0.1            & 34.51$\pm$1.6        & 24.15$\pm$0.6        & 16.19$\pm$0.2         & 51.63$\pm$4.3        & 20.33$\pm$4.9        & 10.60$\pm$1.0        \\
FixMatch     \cite{sohn2020fixmatch}                        & 33.50$\pm$15.1         & 2.56$\pm$0.9          & 2.05$\pm$0.1            & 34.71$\pm$1.4        & 24.48$\pm$0.1        & 16.02$\pm$.1          & 59.87$\pm$3.4        & 22.28$\pm$4.4        &11.59$\pm$1.6        \\
FlexMatch   \cite{zhang2021flexmatch}                         & 29.46$\pm$9.6          & 2.22$\pm$0.3          & 2.12$\pm$0.2            & 36.24$\pm$0.9        & 25.99$\pm$0.5        & 16.28$\pm$.2          & 39.37$\pm$12.9       & 21.83$\pm$3.7        & 10.46$\pm$1.3        \\
Dash     \cite{xu2021dash}                            & 25.65$\pm$4.5          & 3.37$\pm$2.0          & 2.10$\pm$0.3            & 36.67$\pm$0.4        & 25.46$\pm$0.2        & 15.99$\pm$0.2         &58.94$\pm$4.4        & 21.97$\pm$3.9        &10.44$\pm$2.0        \\
AdaMatch      \cite{berthelot2021adamatch}                       & 14.85$\pm$20.4         & 2.06$\pm$0.1          & 2.08$\pm$0.1            & 26.39$\pm$0.1        & 21.41$\pm$0.4        & 15.51$\pm$0.1         & 31.83$\pm$7.7        & 16.50$\pm$4.2        & 10.75$\pm$1.5        \\
InstanT    \cite{li2023instant}                          & \underline{12.68$\pm$10.2}         & 2.07$\pm$0.1          & 1.92$\pm$0.1            & \underline{25.83$\pm$0.3}        & 21.20$\pm$0.4        &15.72$\pm$0.5         & 30.61$\pm$7.4        & 14.91$\pm$2.8        & 10.65$\pm$1.9        \\\hline
InterLUDE (ours)       & 31.90$\pm$4.1   &  \underline{1.78$\pm$0.1}           &    \underline{1.55$\pm$0.1}    &          	35.66$\pm$1.9            &  \underline{21.19$\pm$0.2}         & \underline{13.39$\pm$0.1}    &    	\underline{27.49$\pm$6.6}           &    \textbf{3.14$\pm$0.2}       & \textbf{2.66$\pm$0.1}             \\
InterLUDE+ (ours) & \textbf{12.29$\pm$7.3}   &    \textbf{1.55$\pm$0.1}   & \textbf{1.49$\pm$0.1}    & \textbf{23.60$\pm$1.2}   & \textbf{16.32$\pm$0.3}   & \textbf{12.93$\pm$0.2}  & \textbf{25.83$\pm$9.9} & \underline{4.56$\pm $0.9}  & \underline{3.23$\pm$0.3} \\ \hline
\end{tabular}}
\label{vit-compa}
\end{table*}

\section{Experiments on Heart2Heart Benchmark}
\label{Experiments_OpenSSL}
% In real applications, unlabeled set are often \textit{uncurated}. 
Here we evaluate our method on a open-set medical imaging benchmark proposed in~\citet{huang2023fix}. We compare to strong baselines from Table~\ref{wideresnet-table} as well as two recent SOTA open-set SSL algorithms: OpenMatch~\citep{saito2021openmatch} and Fix-A-Step~\citep{huang2023fix}. 

The Heart2Heart benchmark comprises three fully-deidentified medical image datasets of heart ultrasound images, collected independently by different research groups. It adopts a clinically crucial \textbf{view classification task}: \emph{Given an ultrasound image of the heart, identify the specific anatomical view depicted}. Here we briefly describe the data (Full details in App~\ref{app_dataset_details}). The data for training SSL methods is \textbf{TMED-2} ~\citep{huang2021new,huang2022tmed}, collected in the U.S., $\sim$1700 labeled images (pre-defined train and validation set) of four view types: PLAX, PSAX, A4C and A2C. TMED-2 comes with a large unlabeled set of 353,500 images from 5486 routine scans that are truly \emph{uncurated}, containing out-of-distribution classes, no known true labels, and modest feature distribution shift. The benchmark assesses classifiers on the TMED-2 test set ($\sim$2100 images) as well as the UK-based \textbf{Unity}~\citep{howard2021automated} dataset (7231 images total; PLAX, A2C, and A4C) and France-based \textbf{CAMUS}~\citep{leclerc2019deep} dataset (2000 images total; A2C and A4C).

Heart2Heart poses two key questions: 1. \textbf{Can we train a view classifier from limited labeled data?} (train SSL on TMED-2 then test on TMED-2; Table~\ref{tab:Heart2Heart} column 1) \textbf{2. Can classifiers trained on images from one hospital transfer to hospitals in other countries?} (train SSL on TMED-2 then test on Unity and CAMUS; Table~\ref{tab:Heart2Heart} columns 2-3). 

\textbf{Experiment setting.} 
Our experiments adhere to the exact settings in~\citet{huang2023fix}. We train a separate model for each of the three predefined splits in TMED-2. We use Wide ResNet-28-2 and inherit all common hyperparameters directly from~\citeauthor{huang2023fix} For FreeMatch and FlatMatch, we additionally search $\lambda_{SAF}$ in [0.01, 0.05, 0.1]. For InterLUDE and InterLUDE+, we search $\lambda_{DC}$ in [0.1, 1.0]. We select hyperparameters based on the validation set and report test set performance at the maximum validation checkpoint. Full hyperparameter details are in App.~\ref{sec:app_Heart2Heart_hyperparameters}.

% \textbf{Baselines.}
% We compare our method to several strong closed-set SSL baselines from Table~\ref{wideresnet-table} as well as two recent state-of-the-art open-set SSL methods dedicated to leverage uncurated unlabeled set: OpenMatch~\citep{saito2021openmatch} and FixAStep~\citep{huang2023fix}. 

\begin{table}[h]
\vspace{-.4cm} %% HACK
\caption{Heart2Heart Benchmark. Error rate and standard deviation are reported based on three pre-defined data splits. We re-implemented FlexMatch, FreeMatch and FlatMatch using the author's codes. Other results are copied from~\citet{huang2023fix}. Best results highlighted in bold and the second-best underlined.}
\centering
\begin{adjustbox}{width=0.48\textwidth}\begin{tabular}{l|c|c|c}
\hline
 & TMED2 & CAMUS & UNITY \\ \hline
Supervised         &7.35$\pm$0.79           &30.23$\pm7.77$           & 9.55$\pm$1.68          \\ 
Pi-model         &7.41$\pm$0.63           &38.52$\pm0.96$           &9.90$\pm$1.31           \\ 
VAT         &6.43$\pm$0.11           &32.40$\pm8.11$           &9.21$\pm$2.16           \\ 
FixMatch         &5.66$\pm$0.68           &22.12$\pm9.08$           &7.63$\pm$1.42           \\ 
FlexMatch         &3.56$\pm$0.32           &17.58$\pm6.53$           &\underline{5.20$\pm0.52$}           \\ 
FreeMatch         &3.52$\pm$0.25           &\underline{16.67$\pm6.04$}           &5.45$\pm0.18$           \\ 
FlatMatch         &7.84$\pm$0.48           &28.93$\pm10.38$           &10.05$\pm1.15$           \\ 
FixAStep         &4.79$\pm$0.49           &18.78$\pm10.20$           &6.04$\pm$1.07           \\ 
OpenMatch         &5.88$\pm$0.63           &22.07$\pm5.89$           &7.17$\pm$1.89           \\ \hline
InterLUDE (ours)         &\underline{3.45$\pm$0.39}           &\textbf{13.75$\pm$6.22}           &\textbf{4.86$\pm$0.48}           \\ 
InterLUDE+ (ours)         &\textbf{3.25$\pm$0.17}           &18.12$\pm8.37$           &5.53$\pm0.85$           \\ \hline
\end{tabular}
\end{adjustbox}
\label{tab:Heart2Heart}
\end{table}

\textbf{Results.}
Table~\ref{tab:Heart2Heart} presents our results, with Columns 2 and 3 essentially assessing the \emph{zero-shot cross-hospital generalization} capabilities of each method.
Across all columns, InterLUDE and InterLUDE+ consistently show competitive performance. On TMED2, InterLUDE+ emerges as the top performer, closely followed by InterLUDE. In the zero-shot generalization to CAMUS and Unity, InterLUDE continues to lead, followed by FlexMatch and FreeMatch.
All algorithms show a significant performance drop and increased variance on the CAMUS dataset, a phenomenon also observed in the original paper~\citep{huang2023fix}. 
%which underscores the challenges in this zero-shot generalization evaluations. 
Nevertheless, InterLUDE still outperforms all other baselines, demonstrating strong generalization. FlatMatch on the other hand, substantially underperforms (more than 2x error rate of InterLUDE, more discussion in App~\ref{app_additional_Discussion}).

% FlatMatch underperforms in the Heart2Heart benchmark, possibly because its cross-sharpening objective, aimed at enhancing  ``generalization on unlabeled data'', struggles with TMED2's feature distribution shifts and out-of-distribution classes in the unlabeled set.

% FlatMatch notably underperforms in the Heart2Heart benchmark. We hypothesize that its cross-sharpening objective, motivated by improving ``generalization on unlabeled data'' may fall short in handling the potential feature distribution shift and the presence of out-of-distribution classes in the TMED2's unlabeled set.

% All experiments use the same settings.

\section{Ablations and Sensitivity Analysis}
% \subsection{Ablation}
% \label{ablation}
% Our algorithm contains 2 novel components, the relative loss and the embedding fusion. In table~\ref{tab:CIFAR10_Ablation}, we present an ablation study to measure the contribution of each component. We experiment with both CNN and ViT backbones. More ablation results can be found in App~\ref{app_additional_experiments}

\textbf{Ablation of components.}
\label{ablation}
Our InterLUDE algorithm has two novel components: the cross-instance delta consistency loss and the embedding fusion. Table~\ref{tab:CIFAR10_Ablation} assesses their individual impacts using both CNN and ViT backbones. We find that both components are effective.
On CNN, removing embedding fusion (delta consistency loss) leads to a 0.3\% (0.5\%) drop in performance On ViT, removing embedding fusion (delta consistency loss) leads to performance drops of about 0.5\% (0.6\%).
Note that in this CIFAR-10 40-label scenario, most algorithms compete for improvements of less than 0.5\% over their predecessors, underscoring the effectiveness of both components. More ablations in App Table~\ref{imatch_imatchplus_ViTablation} further confirm the effectiveness of the two components.

\begin{table}[!t]
\vspace{-.4cm} %% HACK
\caption{Ablations to isolate the effect of InterLUDE's two key components. Showing error rate (\%) on CIFAR-10 with 40 labels.}
\centering
\begin{tabular}{l|c|c}
\hline
 & CNN & ViT \\ \hline
InterLUDE         &\textbf{4.51$\pm$0.01} &\textbf{1.78$\pm$0.06}                    \\ 
w/o Embedding Fusion         &4.82$\pm$0.12        &2.31$\pm$0.92            \\
w/o $\mathcal{L}^{\text{DC}}$         &5.00$\pm0.35$   &2.37$\pm0.86$
          \\ \hline
\end{tabular}
\label{tab:CIFAR10_Ablation}
\end{table}

\begin{figure}[h]
  \centering
  \includegraphics[width= 6 cm]{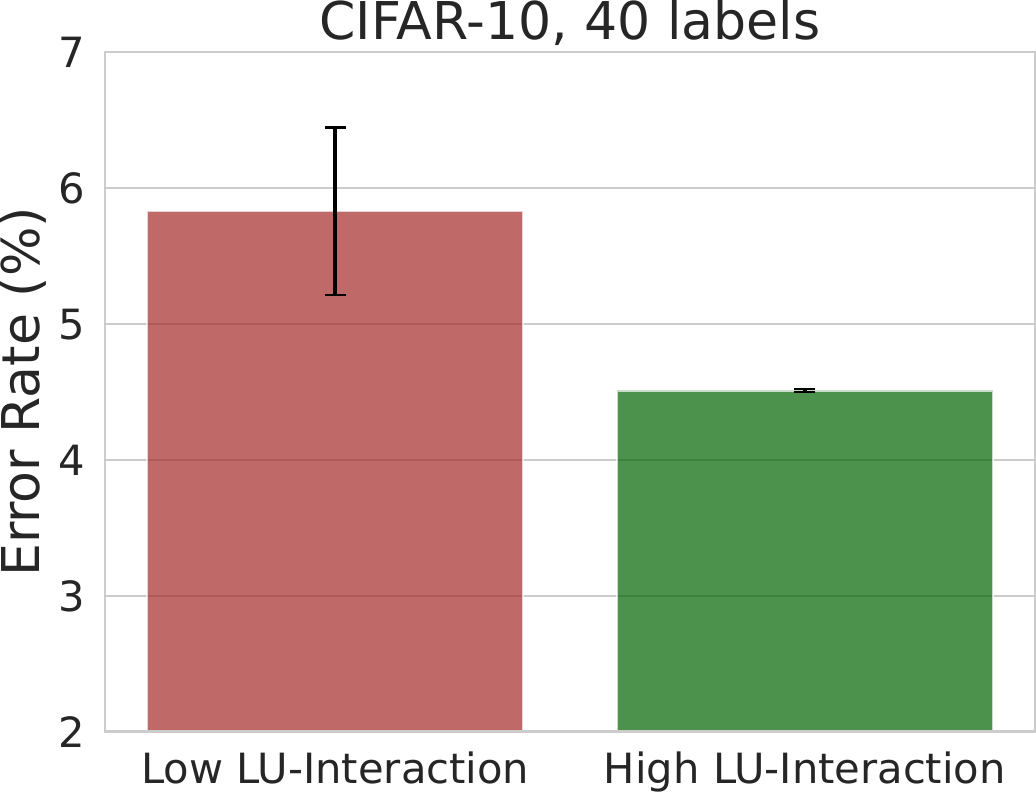}
 % \includegraphics[width= 7 cm]{figs/Ablation_FusionStrategy_CNN_CIFA10_40labels_main.pdf}
%  \vspace{2.0cm}
  \caption{Ablation on Different Batch Layout}
\label{fig:CNN_Ablation_FusionStrategy_CIFAR10_40labels}
\end{figure}

\textbf{Ablation of layout: high vs. low L-U interaction.} Our interdigitated layout (Fig~\ref{embeddingfusion}) is specifically designed to enhance labeled-unlabeled interaction. In Fig~\ref{fig:CNN_Ablation_FusionStrategy_CIFAR10_40labels}, we contrast this layout with a low-interaction alternative that adjacently places all $2B$ labeled augmentations together then all $2\mu B$ unlabeled augmentations together, as follows:
%\text{Low-Interaction}\eqqcolon
\begin{align}
\text{Low-I}\eqqcolon
\{\{x_i^w\}_{i=1}^{B}, \{x_i^{s}\}_{i=1}^{B}, \{\bar{x}_j^w\}_{j=1}^{\mu*B},\{\bar{x}_j^s\}_{j=1}^{\mu*B}\}
\end{align}
Applying circular-shift fusion, the two layout have exactly the same number of within-batch interactions, but our interdigitated layout has far more labeled-unlabeled interactions. Fig~\ref{fig:CNN_Ablation_FusionStrategy_CIFAR10_40labels} shows that our high-LU-interaction interdigitated layout is crucial in the this low label regime. More results can be found in App Fig~\ref{fig:addition_ablation_cifar10_batch_layout}.

% SSL algorithms are often sensitive to hyperparameters~\citep{su2021realistic}. Our framework introduce 2 additional hyperparameters: relative loss coefficient $\lambda_{rel}$ and embedding fusion strength $\alpha$. Here we perform sensitivity analysis on these hyperparameters. 

\textbf{Sensitivity to Delta Consistency Loss Coefficient.} We examined the impact of varying $\lambda_{DC}$, a unique hyperparameter in our method, from 0.1 to 10.0 (see details in App Fig~\ref{AppCNN_Ablation_LambdaREL}). We see stable performance across a wide spectrum of values. However, in extremely low label settings, very high $\lambda_{DC}$ values result in diminished performance, a phenomenon similar to that observed with the unlabeled loss coefficient in other studies~\citep{tarvainen2017mean}.

% \textbf{Sensitivity of Relative Loss Coefficient.} To study the sensitivity of our method on $\lambda_{rel}$, a new hyperparameter our method introduced, we vary $\lambda_{rel}$ from 0.1 to 10.0 (see App Fig~\ref{CNN_Ablation_EmbeddingStrength}). The performance remains relatively stable over a broad range of $\lambda_{rel}$ values. However, in scenarios with extremely limited labels (CIFAR-10 with only 40 labels), excessively high values of $\lambda_{rel}$ lead to performance degradation. This behavior mirrors that commonly seen with the unlabeled losscoefficient~\citep{tarvainen2017mean}.

% \begin{figure}[htb]
%   \centering
%  \includegraphics[width= 8.5cm]{figs/Ablation_LambdaREL_CNN_CIFA10.pdf}
% %  \vspace{2.0cm}
%   \caption{Sensitifity of $\lambda_{rel}$}
%   \label{CNN_Ablation_LambdaREL}
% \end{figure}

\textbf{Sensitivity to Embedding
Fusion strength.} Another hyperparameter introduced by our method is the embedding fusion strength $\alpha$. To evaluate its impact, we vary $\alpha$ from 0.1 to 0.4 (detailed in App Fig~\ref{AppCNN_Ablation_EmbeddingStrength}). We see that $\alpha$ around 0.1 to 0.2 is generally a good value. Excessively high $\alpha$ values, nearing the 0.5 upper limit defined in Eq~\eqref{eq:desiderata} result in a significant drop in performance.

\section{Disccusion}
\label{discussion}
We introduced InterLUDE, an SSL method that fosters deep interactions between labeled and unlabeled data, demonstrating superior performance across multiple SSL benchmarks, particularly with ViT backbones. This underscores the untapped potential of such interactions in SSL. 

% We introduced iMatch, a method that effectively bridges the gap between unlabeled and labeled data through deep interactions in semi-supervised learning (SSL). iMatch demonstrates significant performance superiority over existing competitors across various SSL benchmarks, particularly when utilizing the ViT backbone. Our contention is that deep interactions between labeled and unlabeled data hold substantial potential for SSL applications. Looking ahead, we aim to expand our research into additional SSL scenarios, including domain shift, Robust SSL, and others.

\textbf{Limitations.}
We empirically demonstrate the effectiveness of the embedding fusion and provide some intuition in sec.~\ref{EmbeddingFusion}. More theoretical work is needed to understand the inner mechanism in a more principled way (more discussion in App~\ref{app_additional_discussion_EmbeddingFusion}). Regarding the delta consistency loss, in the extreme case where the regular instance-wise consistency loss leads the model to make \emph{exactly} the same predictions on the weak and strong augmentations for both the unlabeled and labeled data, then the delta consistency loss would approach zero. If this occurs early in the training, the impact of delta consistency loss will be limited. However, extensive empirical evaluations have not observed such an extremity. Further, the current delta consistency loss design relies on average behavior: the deltas of unlabeled data should be \emph{on average} similar to the labeled data. More fine-grained designs can be used, such as enforcing the delta consistency only on instances of the same class (predicted class) across the labeled and unlabeled data.

% in many consistency-regularization based SSL algorithms, the unlabeled loss guides the model to yield similar predictions for the same data under various perturbations. Hypothetically, if this leads to identical predictions for weakly and strongly augmented unlabeled samples, and this invariance extends to labeled data, our proposed relative loss might approach zero. This would limit its impact, especially if it occurs early in training. However, our extensive empirical evaluations have not observed such an extremity, suggesting the relative loss maintains its effectiveness throughout the training process.

% \textbf{Future Work.} In this work, we employ the circular shift as our embedding fusion strategy, with the embedding fusion strength manually determined. Looking ahead, there are two potential directions to further enhance the utilization of embedding fusion in SSL learning tasks. First, as discussed in the paper and appendix, there exists a design space for embedding fusion, allowing exploration of alternative embedding fusion strategies by devising various matrices for $A$. Second, while we manually set the embedding strength and maintain it static throughout the training process, there is an opportunity to explore automatic and dynamic methods for adjusting the embedding fusion strength. Further, more fine-grained control can be applied to the relative loss, such as applying the relative loss on the same class of labeled and unlabeled data (predicted class).

\clearpage
\section{Impact Statement}
% Authors are required to include a statement of the potential broader impact of their work, including its ethical aspects and future societal consequences. This statement should be in a separate section at the end of the paper (co-located with Acknowledgements, before References), and does not count toward the paper page limit. In many cases, where the ethical impacts and expected societal implications are those that are well established when advancing the field of Machine Learning, substantial discussion is not required, and a simple statement such as: 

This paper presents work whose goal is to advance the field of semi-supervised learning.
We hope our work improves practitioners' ability to train accurate classifiers from limited labeled data, especially in medical applications where acquiring labeled data is prohibitively costly. Even though the Heart2Heart cross-hospital generalization task we examine is fully deidentified and open-access by the respective dataset creators, we implore future work pursuing it to remember the real human patients the data represents and take proper care. A key concern in translating SSL-trained classifiers to practice is fairness to different subpopulations, which cannot yet be assessed with the available data in that benchmark.

%Learning. There are many potential societal consequences of our work, none of which we feel must be specifically highlighted here.

% The above statement can be used verbatim in such cases, but we encourage authors to think about whether there is content which does warrant further discussion, as this statement will be apparent if the paper is later flagged for ethics review.}

% % Acknowledgements should only appear in the accepted version.
% \section*{Acknowledgements}

\renewcommand{\bibsection}{\subsubsection*{References}}
\bibliographystyle{abbrvnat}
\bibliography{main.bib}


\newpage
\appendix
\onecolumn

\begin{center}
\Large Supplementary Material
\end{center}
~\\
~\\
In this supplement, we provide:
\begin{itemize}
  \setlength\itemsep{0em}
\item Sec.~\ref{app_dataset_details}: Dataset Details
\item Sec.~\ref{app_hyperparameter_details}: Hyperparameter Details

\item Sec.~\ref{app_iMatch+_details}: InterLUDE+ Details

\item Sec.~\ref{app_additional_experiments}: Additional Albation and Sensitivity Analysis

\item Sec.~\ref{app_additional_Discussion}:  Additional Discussion
\end{itemize}

\section{Additional Dataset Details}
\label{app_dataset_details}
Here, we provide more details on the datasets used in the paper. For the medical image datasets used in the Heart2Heart Benchmark~\citep{huang2023fix}, we emphasize that all three medical image datasets are deidentified and accessible to academic researchers. 

\subsection{Classic Benchmarks}
\begin{itemize}
    \item The CIFAR-10 dataset is a collection of images commonly used to train machine learning and computer vision algorithms, and has been a classic benchmark to use for SSL. It contains 60,000 32x32 color images in 10 different classes, with 6,000 images per class. The dataset is divided into 50,000 training images and 10,000 testing images. The classes include various objects and animals like cars, birds, dogs, and ships.. 
    
    \item Similar to CIFAR-10, the CIFAR-100 dataset is another common SSL benchmark. CIFAR-100 has same total number of images as CIFAR-10, i.e., 60,000 32x32 color images, but they are spread over 100 classes, each containing 600 images, with 500 training images and 100 testing images per class. The dataset features a more diverse set of classes compared to CIFAR-10
    
    \item The STL-10 dataset is also popular in benchmarking SSL algorithms. The dataset contains 5,000 labeled training images and 8,000 test images, with each image being a higher resolution of 96x96 pixels. It contains 10 classes: airplane, bird, car, cat, deer, dog, horse, monkey, ship, truck. Additionally, the dataset provides 100,000 unlabeled images for unsupervised learning, which are drawn from a similar but broader distribution than the labeled images.

\end{itemize}

\subsection{Heart2Heart Benchmark}
The Heart2Heart Benchmark contains three medical image
datasets of heart ultrasound images, collected independently
by different research groups in the world. Thanks to common device standards, these images are interoperable. All data used in the benchmark are resized to 112x112~\citep{huang2023fix}.

The benchmark adopts a view classification task: \emph{Given an ultrasound image of the heart, identify the specific anatomical view depicted}. Such task is of great clinical importance, as determining the view type is a prerequisite for many clinical measurements and diagnoses~\citep{madani2018fast}.

\begin{itemize}
    \item The TMED-2 dataset provides set of labeled images of four specific view types: Parasternal Long Axis (PLAX), Parasternal Short Axis (PSAX), Apical Four Chamber (A4C) and Apical Two Chamber (A2C), gathered from certified annotators, and a truly \emph{uncurated} unlabeled set of 353,500 images from routine scans of 5486 patient-studies. With routine Transthoracic Echocardiograms (TTEs) commonly presenting at least nine canonical view types~\citep{mitchell2019guidelines}, this unlabeled set likely contains classes not found in the labeled set. 
    Moreover, the unlabeled dataset consists of a wide variety of patient scans, in contrast to the labeled dataset, which specifically contains a higher percentage of patients with a heart disease called aortic stenosis (AS), particularly severe AS. About 50\% of patients in the labeled set have severe AS, a significant increase compared to its less than 10\% occurrence in the general population. This discrepancy in sampling results in noticeable feature differences. For instance, PLAX and PSAX images from patients with severe AS typically exhibit more pronounced calcification (thickening) of the aortic valve.

    Note that the TMED-2 data used in Heart2Heart Benchmark is only \textbf{a subset of the full TMED-2 dataset}, more details should be referred to the original papers~\citep{huang2022tmed,huang2021new}
    
    \item The UNITY dataset contains ultrasound images of the heart collected from 17 hospitals in the UK. The original dataset contains other view such as A3C and A5C, but only PLAX, A2C and A4C are used in the Heart2Heart Benchmark. More details should be referred to the original paper~\citep{howard2021automated}.

    \item The CAMUS dataset contains ultrasound images of the heart collected from a hospital in France. The dataset contains A2C and A4C. Beside the view label, the original dataset also provide labels for the cardiac cycle when the image is acquired (e.g., end diastolic or end systolic).  More details should be referred to the original paper~\citep{leclerc2019deep}.

\end{itemize}

\section{Additional Hyperparameter Details}
\label{app_hyperparameter_details}
Here we provide complete list of the hyperparameters used in the paper:

\subsection{Classic SSL Benchmarks}
\textbf{CNN Backbones.} Here we list the hyperparameters used for CNN backbone experiments in Sec~\ref{Experiments}. We closely follow the established setting from prior studies~\citep{wang2022freematch}. 
\begin{table}[ht]
\centering
\caption{Algorithm-independent hyperparameters on CNNs}
\label{tab:hyperparameters_ClassicBenchmark_CNN}
\begin{tabular}{lccc}
\toprule
Dataset        & CIFAR-10   & CIFAR-100  & STL-10 \\
\midrule
Model          & WRN-28-2   & WRN-28-8   & WRN-37-2   \\
Weight decay   & 5e-4       & 1e-3       & 5e-4       \\
Batch size     & 64         & 64         & 64        \\
Learning rate  & 0.03       & 0.03       & 0.03       \\
SGD momentum   & 0.9        & 0.9        & 0.9        \\
EMA decay      & 0.999      & 0.999      & 0.999      \\
\bottomrule
\end{tabular}
\end{table}

For algorithm dependent hyperparameters, we set $\lambda_{DC}$ to 1.0, embedding fusion strength $\alpha$ to 0.1. We set the unlabeled batch ratio $\mu$ to 7 following convention~\citep{sohn2020fixmatch,zheng2023simmatchv2,wang2022freematch}. Note that $\mu$ is a common hyperparameter to many SSL algorithms, not unique to ours.

\textbf{ViT Backbones.} Here we list the hyperparameters used for the ViT backbone experiments in Sec~\ref{Experiments}. We closely follow the established setting from prior studies~\citep{li2023instant}. For the ViT-small experiments, we utilized the pretrained weights from ViT-small as provided by \cite{Wang2022USB}. In the case of ViT-base experiments, we employed the pretrained weights from PyTorch image models, as documented in \cite{Imambi2021Pytorch}.
\begin{table}[ht]
\centering
\caption{Algorithm-independent hyperparameters on ViTs}
\label{tab:hyperparameters_ClassicBenchmark_ViT}
\begin{tabular}{lccc}
\toprule
Dataset        & CIFAR-10   & CIFAR-100  & STL-10 \\
\midrule
Model          & ViT-small   & ViT-small    & ViT-base   \\
Weight decay   & 5e-4        & 5e-4       & 5e-4      \\
Batch size     & 8         & 8         & 8        \\
Learning rate  & 5e-4      & 5e-4      &  1e-4       \\
Optimizer   & AdamW        & AdamW        & AdamW        \\
EMA decay      & 0.999      & 0.999      & 0.999     \\
\bottomrule
\end{tabular}
\end{table}

For algorithm dependent hyperparameters, we set $\lambda_{DC}$ to 0.1, embedding fusion strength $\alpha$ to 0.1. We set the unlabeled batch ratio $\mu$ to 7 following convention~\citep{sohn2020fixmatch,zheng2023simmatchv2,wang2022freematch}. Note that $\mu$ is a common hyperparameter to many SSL algorithms, not unique to ours.

\subsection{Heart2Heart Benchmark}
\label{sec:app_Heart2Heart_hyperparameters}
Our experiments closely follow the protocol from~\citep{huang2023fix}. We directly inherit all the common hyperparameters from~\citep{huang2023fix} without tunning. For FreeMatch and FlatMatch, we additionally search $\lambda_{SAF}$ in [0.01, 0.05, 0.1]. For InterLUDE and InterLUDE+ we search $\lambda_{DC}$ in [0.1, 1.0].

\begin{table}[ht]
\centering
\caption{Heart2Heart Benchmark Hyperparameters}
\label{tab:hyperparameters_Heart2Heart}
\begin{tabular}{lccc}
\toprule
        & Data Split 1   & Data Split 2 & Data Split 3 \\
\hline

Weight decay   & 5e-4        & 5e-4       & 5e-4      \\
Learning rate  & 0.1      & 0.1      &  0.1       \\
Batch size     & 64         & 64         & 64        \\
Optimizer   & SGD        & SGD        & SGD        \\
\hline
\multicolumn{4}{c}{FreeMatch} \\
\hline
$\lambda_{SAF}$ & 0.05 & 0.01 & 0.05\\
\hline
\multicolumn{4}{c}{FlatMatch} \\
\hline
$\lambda_{SAF}$ & 0.05 & 0.1 & 0.01\\
\hline
\multicolumn{4}{c}{InterLUDE} \\
\hline
$\lambda_{DC}$ & 1.0 & 0.1 & 0.1\\
\hline
\multicolumn{4}{c}{InterLUDE+} \\
\hline
$\lambda_{DC}$ & 1.0 & 0.1 & 0.1\\
\bottomrule
\end{tabular}
\end{table}

\section{InterLUDE+ Details}
\label{app_iMatch+_details}

Recently, Self-Adaptive Threshold (SAT) and Self-Adaptive Fairness (SAF) were introduced by~\citet{wang2022freematch}, with potential applicability to other SSL algorithms. Here, we integrate SAT and SAF into our InterLUDE framework, and named the enhanced version InterLUDE+. 

SAT aim to adjust the pseudo-label threshold by considering both the global (dataset-specific) threshold and local (class-specific) threshold at different time step, each estimated via the model's current learning status. Similar techniques such as Distribution Alignment (DA)~\citep{berthelot2019remixmatch} and its variant~\citep{berthelot2021adamatch} has been proposed and used by various SSL algorithms~\citep{li2021comatch,zheng2022simmatch}. 
\begin{align}
\tau_t(c) &= \frac{\tilde{p}_t(c)}{\max\{\tilde{p}_t(c) : c \in [C]\}} \cdot \tau_t,
\end{align}
where $\tau_{t}$ is the gloabl threshold at time $t$ and $\tilde{p}_t(c)$ is the local threshold for class $c$ at time $t$.
\begin{align}
\tau_t =
\begin{cases} 
\frac{1}{C}, & \text{if } t = 0,\\
\lambda \tau_{t-1} + (1 - \lambda) \frac{1}{\mu B}
\sum_{j=1}^{\mu B}
\max(q_{j}^{w}), & \text{otherwise},
\end{cases}
\end{align}

\begin{align}
\tilde{p}_t(c) = 
\begin{cases} 
\frac{1}{C}, & \text{if } t = 0,\\
\lambda \tilde{p}_{t-1}(c) + (1 - \lambda) \frac{1}{\mu B} \sum_{j=1}^{\mu B} q_{j}^{w}(c), & \text{otherwise},
\end{cases}
\end{align}

SAF is a regularization term that encourages diverse model predictions on unlabeled data.

\begin{align}
\mathcal{L}^{SAF} &= -H \left[ \text{SumNorm} \left( \frac{\tilde{p}_t}{\tilde{h}_t} \right), \text{SumNorm} \left( \frac{\bar{p}}{\bar{h}} \right) \right]
\end{align}

where 
\begin{align}
\bar{p} &= \frac{1}{\mu B} \sum_{j=1}^{\mu B} \mathbf{1} (\max (q_{j}^w) \geq \tau_t (\arg\max (q_{j}^w)) q_{j}^s, \\
\bar{h} &= \text{Hist}_{\mu B} \left( \mathbf{1} (\max (q_{j}^w) \geq \tau_t (\arg\max (q_{j}^w)) \hat{q_{j}^s} \right),
\\
\tilde{h}_t &= \lambda \tilde{h}_{t-1} + (1 - \lambda) \text{Hist}_{\mu B} (\hat{q_{j}^w})
\end{align}

$\text{SumNorm} = (\cdot) / \sum (\cdot)$. Hist means the histogram distribution. $\hat{q_{j}^w}$ and $\hat{q_{j}^s}$ are the one-hot encoding of $q_{j}^w$ and $q_{j}^s$. More details of the SAT and SAF should be referred to in the original paper~\citep{wang2022freematch}.

Incorporating SAT and SAF with InterLUDE, the loss function of InterLUDE+ can be written as: 
\begin{align}
\mathcal{L} &= \mathcal{L}^{L} + \lambda_{u}\mathcal{L}^{U} + \lambda_{DC} \mathcal{L}^{DC} + \lambda_{SAF} \mathcal{L}^{SAF}\label{APPeq:imatchplus_total_loss}
\end{align}

where $\tau_t(c)$ is used to replaced the fixed threshold $\tau$ in Eq.~\eqref{eq:instance_wise_unlabeled_loss}.

\section{Additional Experiments}
\label{app_additional_experiments}

\subsection{CNN backbones}
\subsubsection{Sensitivity Analysis}
\begin{figure}[htb]
  \centering
  \begin{minipage}{0.48\textwidth}
    \centering
    \includegraphics[width=\linewidth]{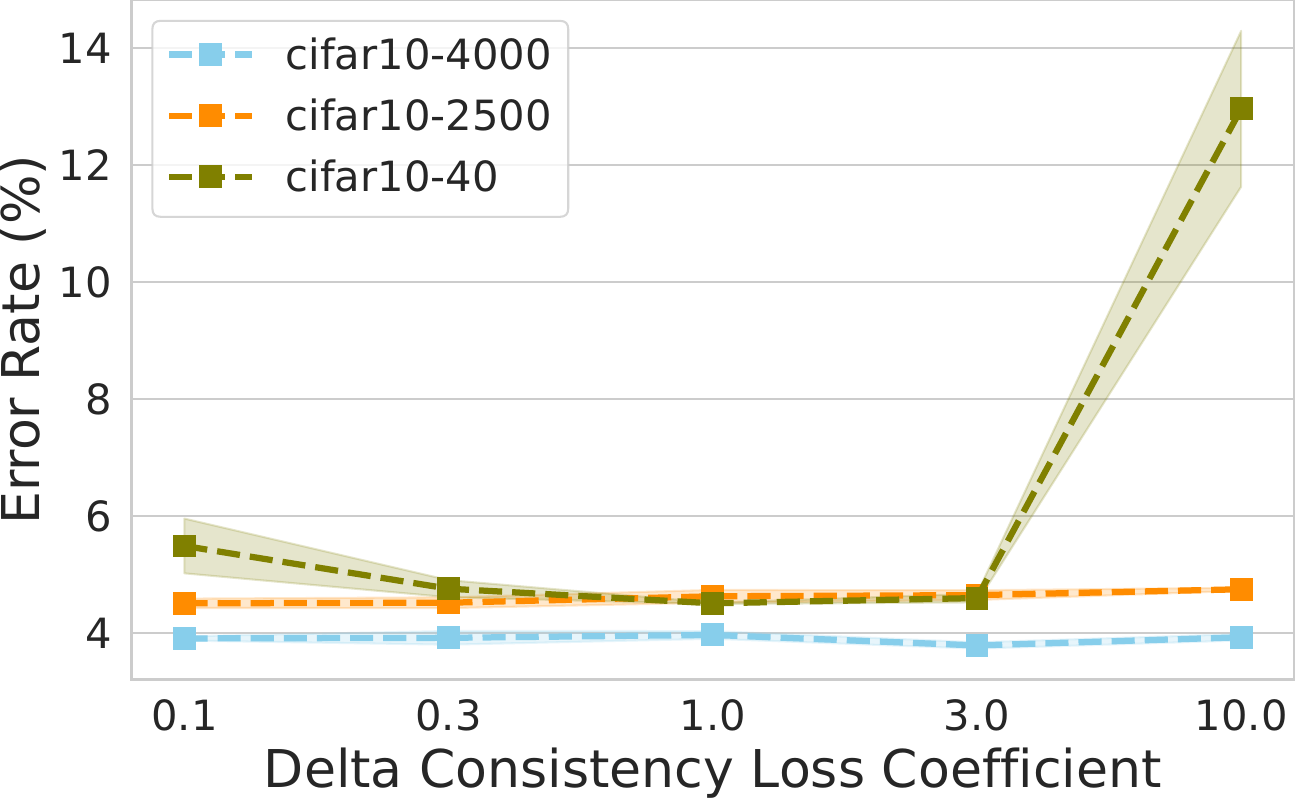}
    \caption{Sensitivity of Delta Consistency Loss Coefficient. Showing CNN performance on CIFAR-10.}
    \label{AppCNN_Ablation_LambdaREL}
  \end{minipage}\hfill
  \begin{minipage}{0.48\textwidth}
    \centering
    \includegraphics[width=\linewidth]{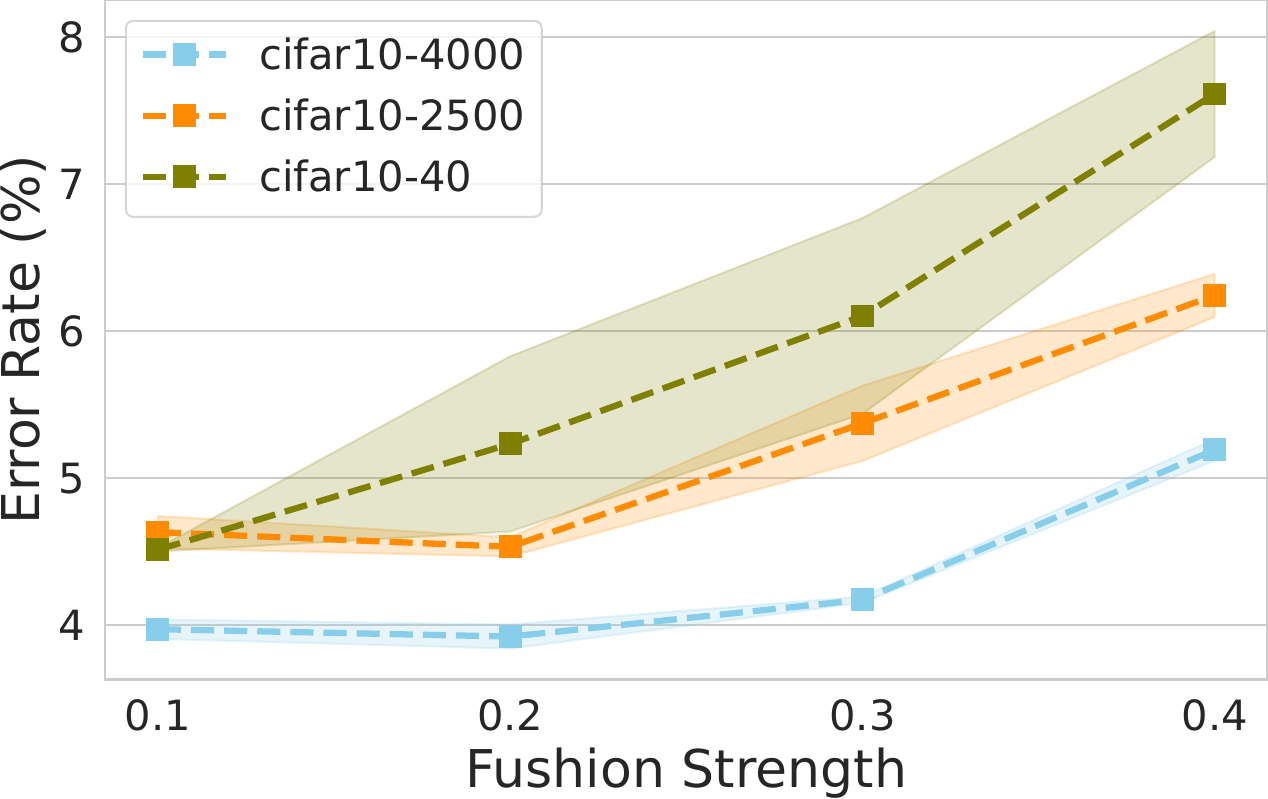}
    \caption{Sensitivity of Embedding Fusion Strength. Showing CNN performance on CIFAR-10.}
    \label{AppCNN_Ablation_EmbeddingStrength}
  \end{minipage}
\end{figure}

The sensitivity analysis of hyperparameters unique to our method: delta consistnecy loss $\lambda_{DC}$ and embedding fusion strength $\alpha$ with CNN backbone on the CIFAR10 dataset are presented in Fig. \ref{AppCNN_Ablation_LambdaREL} and \ref{AppCNN_Ablation_EmbeddingStrength}, respectively. We can observe that the error remains stable over a wide range of $\lambda_{DC}$, leading us to conclude that InterLUDE is not overly sensitive to $\lambda_{DC}$. When considering the embedding fusion strength, we see that $\alpha$ around 0.1 to 0.2 is generally a good value.   Excessively high $\alpha$ values, nearing the 0.5 upper limit defined in Eq.~\eqref{eq:desiderata} result in a significant drop in performance. More sensitivity analysis on the ViT backbones can be found in sec.~\ref{app_ViT_sensitivity_analysis}.

\subsubsection{Additional Ablation of layout: high vs. low L-U interaction.}

Our interdigitated batch layout (Fig~\ref{embeddingfusion}) is specifically designed to enhance labeled-unlabeled interaction. Here we conduct more ablations to analysis the effect of high vs. low L-U interaction under the same circular shift embedding fusion operation. We contrast the low L-U interaction batch layout with 3 high L-U interaction layouts (including the one we used as default in the paper: Fig~\ref{embeddingfusion})

The different layouts are defined as follows:
%\text{Low-Interaction}\eqqcolon
\begin{align}
\text{Low-I}&\eqqcolon
\{\{x_i^w\}_{i=1}^{B}, \{x_i^{s}\}_{i=1}^{B}, \{\bar{x}_j^w\}_{j=1}^{\mu*B},\{\bar{x}_j^s\}_{j=1}^{\mu*B}\}\\
\text{High-I1}&\eqqcolon\{\{x_i^w\}_{i=1}^{B/2(\mu+1)}, \{x_i^{s}\}_{i=1}^{B/2(\mu+1)}, \{\bar{x}_j^w\}_{j=1}^{(\mu*B)/2(\mu+1)},\{\bar{x}_j^s\}_{j=1}^{(\mu*B)/2(\mu+1)}\}^{2(\mu+1)}\\
\text{High-I2}&\eqqcolon
\{\{x_i^w\}, \{x_i^{s}\}, \{\bar{x}_j^w\}_{j=1}^{\mu},\{\bar{x}_j^s\}_{j=1}^{\mu}\}^{B}\\
\text{High-I3 (default)}&\eqqcolon
\{\{x_i^w\}, \{\bar{x}_j^w\}_{j=1}^{\mu},\{x_i^{s}\}, \{\bar{x}_j^s\}_{j=1}^{\mu}\}^{B}\\
\end{align}

Note that applying the same circular-shift fusion, these different  layout have exactly the same number of within-batch interactions, the high-L-U-interaction design has more labeled-unlabeled interactions. 

\begin{figure}[htb]
  \centering
  \begin{minipage}{0.32\textwidth}
    \includegraphics[width=\linewidth]{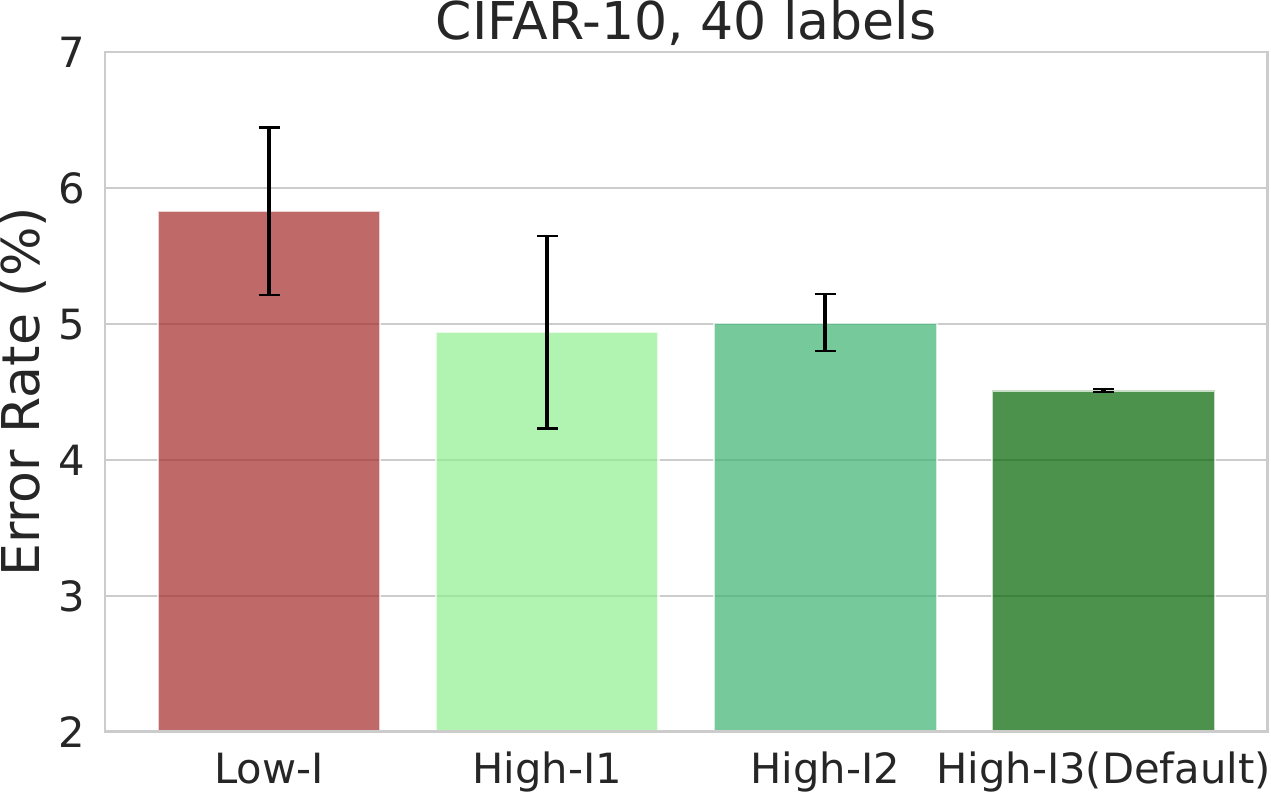}
    % \caption{Different Batch Layout, CIFAR10 40 labels}
    \label{fig:cifar10_40}
  \end{minipage}
  \hfill
  \begin{minipage}{0.32\textwidth}
    \includegraphics[width=\linewidth]{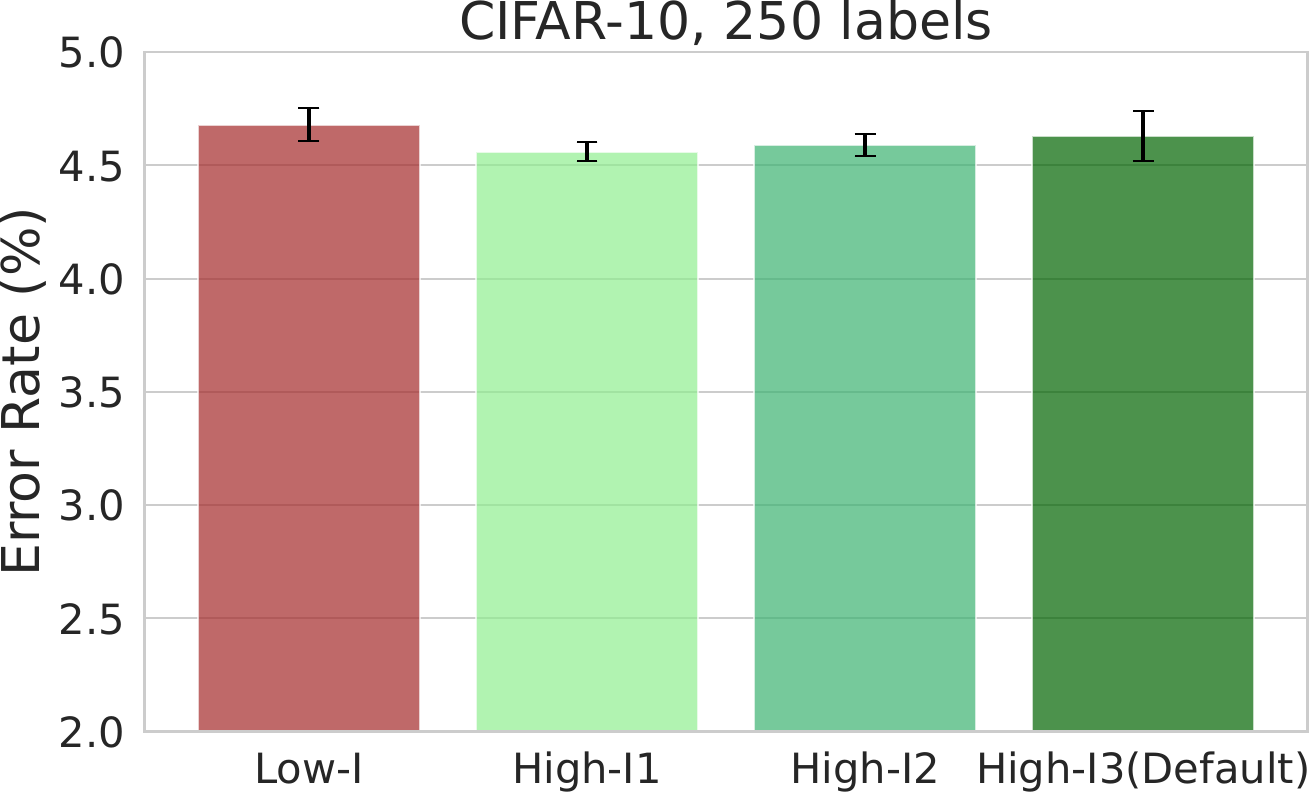}
    % \caption{Effect of Batch Layout on CIFAR10 250 labels}
    \label{fig:cifar10_250}
  \end{minipage}
  \hfill
  \begin{minipage}{0.32\textwidth}
    \includegraphics[width=\linewidth]{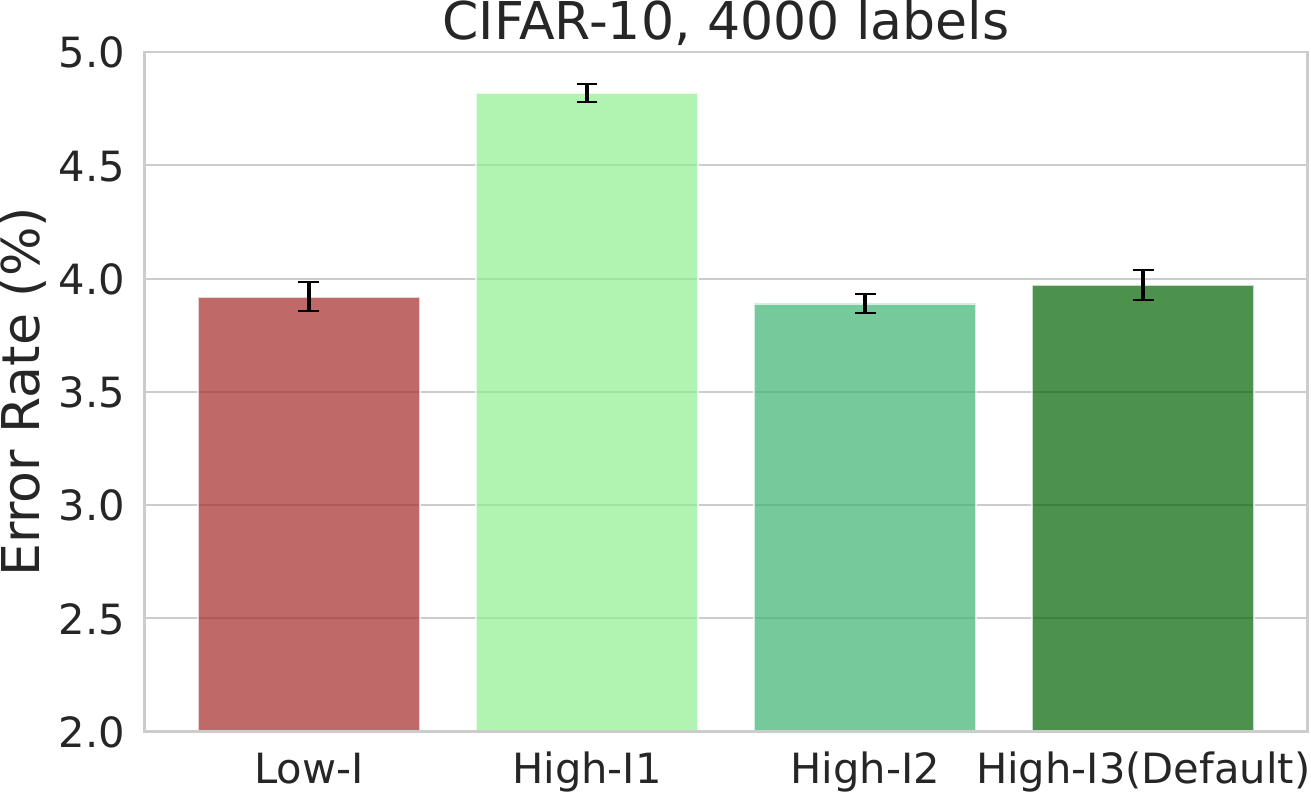}
    % \caption{Different Batch Layout, CIFAR10 4000 labels}
    \label{fig:cifar10_4000}
  \end{minipage}
  \caption{Effect of Different Batch Layout. Showing result on CIFAR-10. From left to right are CIFAR-10 with 40 labels, 250 labels and 4000 labels. }
  \label{fig:addition_ablation_cifar10_batch_layout}
\end{figure}

We can see from Fig~\ref{fig:addition_ablation_cifar10_batch_layout} that having more L-U interactions is \emph{especially important in the low label regime}. However, as the labeled examples becoming more available (e.g., CIFAR-10 with 4000 labels scenario), the advantage of having L-U interactions over L-L interactions diminishes, which is intuitive: \emph{if we have enough labeled data, we might not need the unlabeled data}. 

% $\text{Low-I}\eqqcolon\{l_{weak}*B, l_{strong}*B, u_{weak}*\mu*B, u_{strong}*\mu*B\}$

% $\text{High-I1}\eqqcolon\{l_{weak}*(B/2(\mu+1)), l_{strong}*(B/2(\mu+1)), u_{weak}*((\mu*B)/2(\mu+1)), u_{strong}*((\mu*B)/2(\mu+1))\}*2(\mu+1)$

% $\text{High-I2}\eqqcolon\{l_{weak}, l_{strong}, u_{weak}*\mu, u_{strong}*\mu\}*B$

% $\text{High-I3(default)}\eqqcolon\{l_{weak}, u_{weak}*\mu, l_{strong}, u_{strong}*\mu\}*B$

\subsubsection{Wall Time Comparison}
\label{sec:WallTimeComparison_TMED2}
Here we compare the wall time of FlatMatch~\citep{huang2023flatmatch} and InterLUDE on the TMED2 dataset using the \textbf{exact same hardware} (NVIDIA A100 with 80G Memory). We can see that InterLUDE has a substantially lower run time cost per iteration.
\begin{figure}[htb]
  \centering
 \includegraphics[width= 8.5cm]{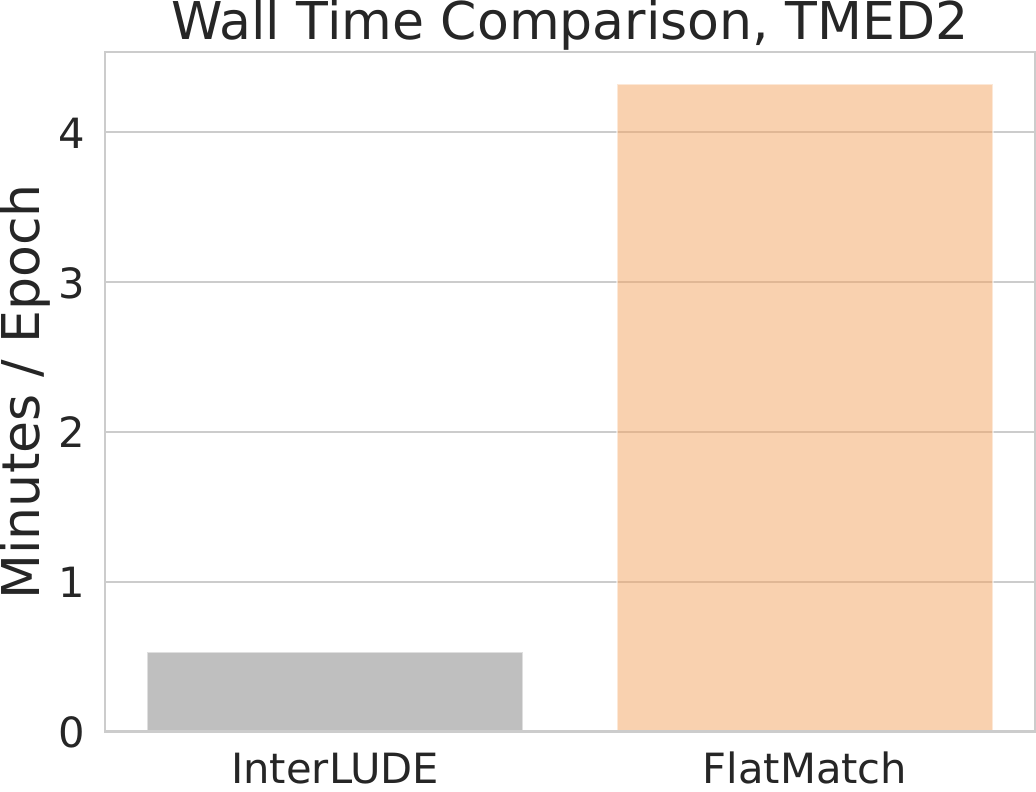}
%  \vspace{2.0cm}
  \caption{Wall Time Comparison}
\end{figure}

% \begin{figure}[ht]
%   \centering
%   \begin{minipage}{0.48\textwidth}
%     \centering
%     \includegraphics[width=\linewidth]{figs/Ablation_LambdaREL_ViT_CIFA10_Label10_ErrorRate.pdf}
%     \caption{Sensitivity of Delta Consistency Loss. Showing ViT performance on 10 labels of CIFAR-10.}
%     \label{AppViT_Ablation_LambdaREL_cifar10_10}
%   \end{minipage}\hfill
%   \begin{minipage}{0.48\textwidth}
%     \centering
%     \includegraphics[width=\linewidth]{figs/Ablation_LambdaREL_ViT_CIFA10_Label40250_ErrorRate.pdf}
%     \caption{Sensitivity of Delta Consistency Loss Coefficient. Showing ViT performance on 40 and 250 labels of CIFAR-10.}
%     \label{AppViT_Ablation_LambdaREL_cifar10_40_250}
%   \end{minipage}
%   \begin{minipage}{0.48\textwidth}
%     \centering
%     \includegraphics[width=\linewidth]{figs/Ablation_PN_ViT_CIFA10_Label10_ErrorRate.pdf}
%     \caption{Sensitivity of Embedding Strength. Showing ViT performance on 10 labels of CIFAR-10.}
% \label{AppViT_Ablation_EmbeddingStrength_cifar10_10}
%   \end{minipage}\hfill
%   \begin{minipage}{0.48\textwidth}
%     \centering
%     \includegraphics[width=\linewidth]{figs/Ablation_PN_ViT_CIFA10_Label40250_ErrorRate.pdf}
%     \caption{Sensitivity of Embedding Strength. Showing ViT performance on 40 and 250 labels of CIFAR-10.}
%     \label{AppViT_Ablation_EmbeddingStrength_cifar10_40_250}
%   \end{minipage}
% \end{figure}

\subsection{ViT backbones}
\subsubsection{Ablation}
\label{sec:vit_ablation}
The ablation of InterLUDE and InterLUDE$+$ with ViT backbone are presented in Table \ref{imatch_imatchplus_ViTablation}. We can observe that the delta consistency loss and embedding fusion both contribute to the performance improvement of the proposed method.

\begin{table*}[htb]
\caption{Ablation study on ViT backbone.   Error rates (\%)  are averaged with three random seeds and reported with a 95\% confidence interval. }
\resizebox{\textwidth}{!}{\begin{tabular}{c|cc|cc|cc}
\hline
Dataset                        & \multicolumn{2}{c|}{CIFAR10} & \multicolumn{2}{c|}{CIFAR100} & \multicolumn{2}{c}{STL10}  \\ \hline
\#Label                        & 40            & 250          & 400           & 2500          & 40           & 100          \\ \hline
InterLUDE                         & 1.78$\pm$0.1  & 1.55$\pm$0.1 & 21.19$\pm$0.2 & 13.39$\pm$0.1 & 3.14$\pm$0.2 & 2.66$\pm$0.1 \\
InterLUDE (w/o Embedding Fusion)  & 2.31$\pm$0.9  & 1.64$\pm$0.1 & 23.70$\pm$1.9 & 13.88$\pm$0.3 & 5.23$\pm$3.7 & 3.23$\pm$0.4 \\
InterLUDE (w/o $\mathcal{L}^{\text{DC}}$)     & 2.37$\pm$0.9  & 1.75$\pm$0.2 & 23.20$\pm$1.7 & 14.07$\pm$0.6 & 3.75$\pm$0.4 & 3.36$\pm$0.3 \\ \hline
InterLUDE+                        & 1.55$\pm$0.1  & 1.49$\pm$0.1 & 16.32$\pm$0.3 & 12.93$\pm$0.2 & 4.56$\pm$0.9 & 3.23$\pm$0.3 \\
InterLUDE+ (w/o Embedding Fusion) & 1.78$\pm$0.1  & 1.61$\pm$0.1 & 16.63$\pm$1.6 & 14.00$\pm$0.6 & 4.25$\pm$1.0 & 4.37$\pm$0.1 \\
InterLUDE+ (w/o $\mathcal{L}^{\text{DC}}$)        & 1.57$\pm$0.1  & 1.60$\pm$0.1 & 16.33$\pm$0.8 & 13.15$\pm$0.1 & 4.71$\pm$1.0 & 4.55$\pm$0.3 \\ \hline
\end{tabular}}
\label{imatch_imatchplus_ViTablation}
\end{table*}

\subsubsection{Sensitivity Analysis}
\label{app_ViT_sensitivity_analysis}
\begin{figure}[htb]
  \centering
  \begin{minipage}{0.48\textwidth}
    \centering
    \includegraphics[width=\linewidth]{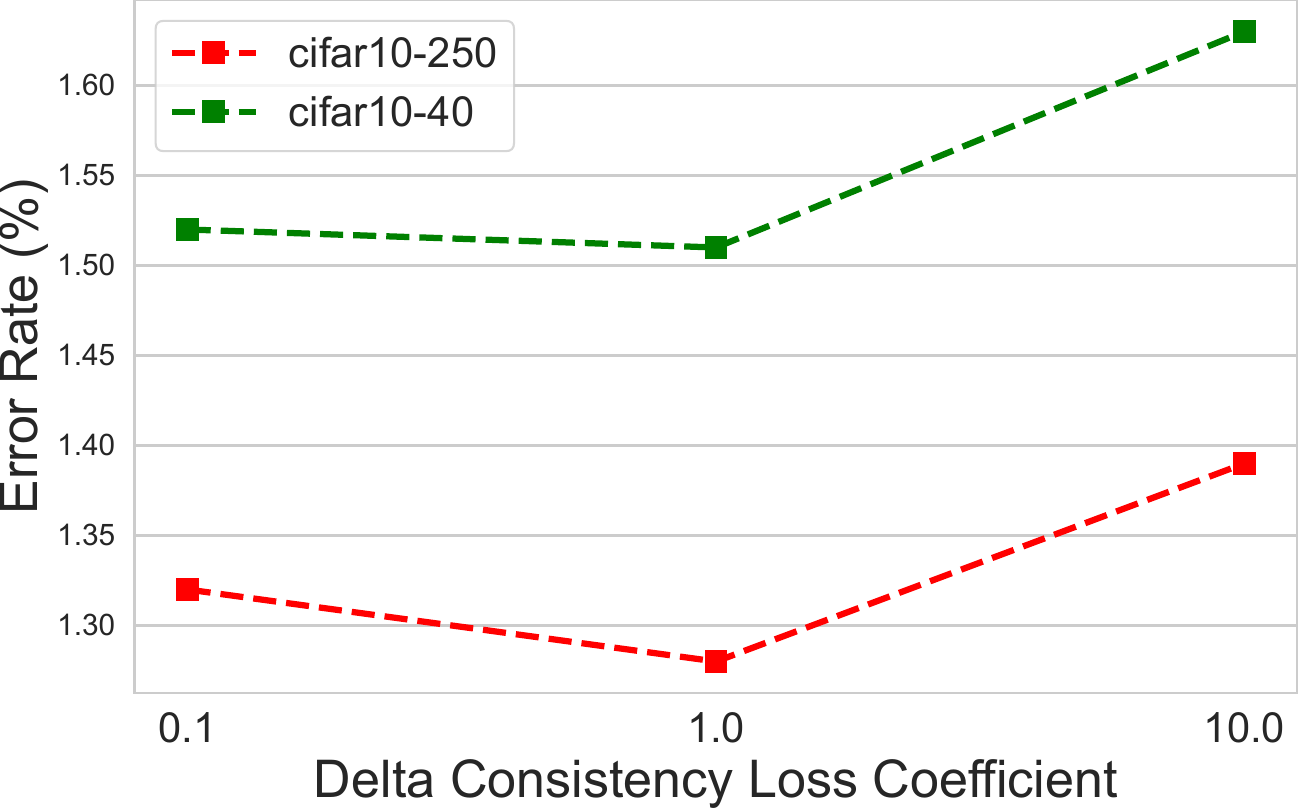}
    \caption{Sensitivity of Delta Consistency Loss Coefficient. Showing ViT performance on CIFAR-10.}
    \label{AppViT_Ablation_LambdaREL_cifar10_40_250}
  \end{minipage}\hfill
  \begin{minipage}{0.48\textwidth}
    \centering
    \includegraphics[width=\linewidth]{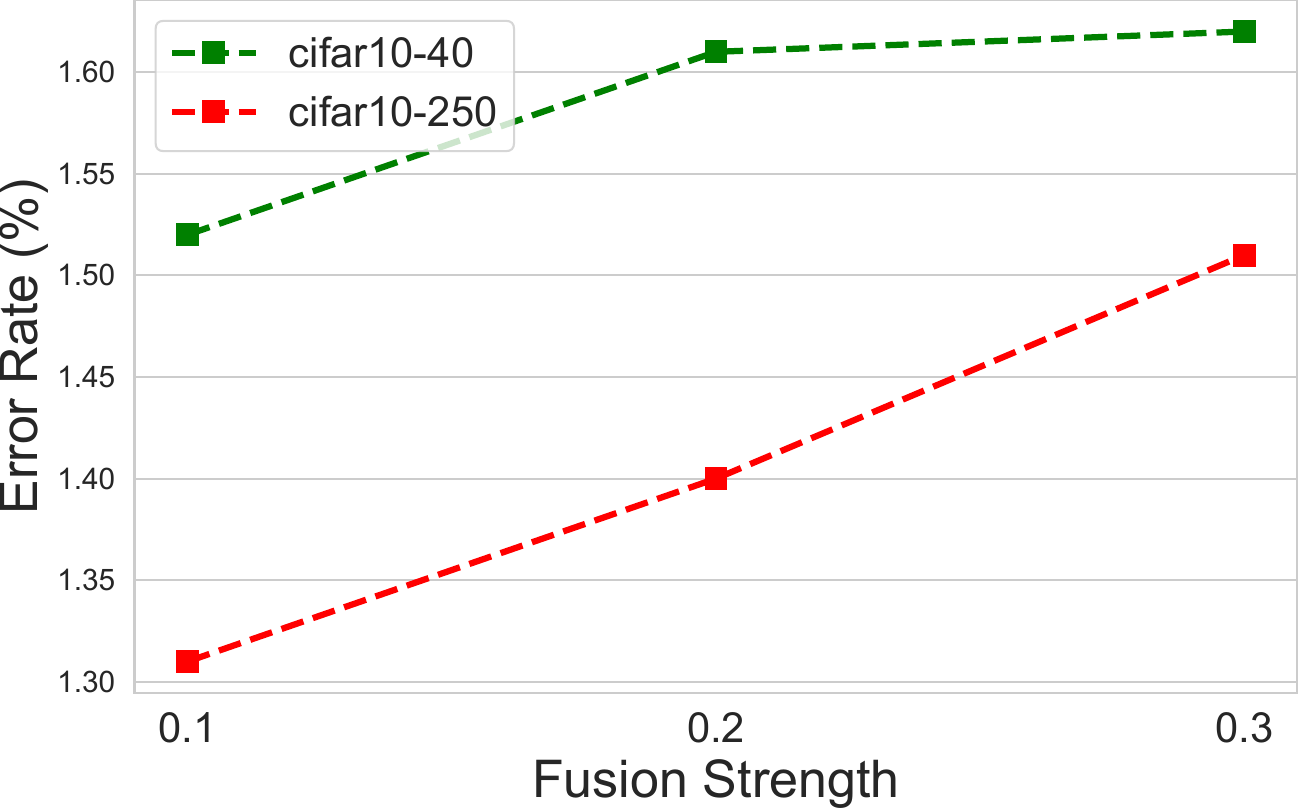}
    \caption{Sensitivity of Embedding Fusion Strength. Showing ViT performance on CIFAR-10.}
    \label{AppViT_Ablation_EmbeddingStrength_cifar10_40_250}
  \end{minipage}
\end{figure}

Due to limited computational resources, we conduct a single experiment for sensitivity analysis on the ViT backbone. The sensitivity analysis of the hyperparameters of the delta consistency loss $\lambda_{DC}$ and embedding fusion strength $\alpha$ with ViT backbone on the CIFAR10 dataset are presented in Fig.  \ref{AppViT_Ablation_LambdaREL_cifar10_40_250}, and \ref{AppViT_Ablation_EmbeddingStrength_cifar10_40_250} respectively. \iffalse We present the settings with 10 labels separately from those with 40/250 labels for better visualization, owing to the substantial scale difference between the 10-label and 40/250-label settings.\fi

% \iffalse
% \begin{figure}[htb]
%   \centering
%  \includegraphics[width= 8.5cm]{figs/Ablation_LambdaREL_ViT_CIFA10.pdf}
% %  \vspace{2.0cm}
%   \label{FusionStrategy}
%   \caption{Abalation on LambdaREL. ViT backbone.}
% \end{figure}
% \begin{figure}[htb]
%   \centering
%  \includegraphics[width= 8.5cm]{figs/Ablation_LambdaREL_ViT_CIFA10_40_250.pdf}
% %  \vspace{2.0cm}
%   \label{FusionStrategy}
%   \caption{Abalation on LambdaREL. ViT backbone.}
% \end{figure}
% \fi

% \begin{figure}[htb]
%   \centering
%   \begin{minipage}{0.48\textwidth}
%     \centering
%     \includegraphics[width=\linewidth]{figs/Ablation_PN_ViT_CIFA10_Label10_ErrorRate.pdf}
%     \caption{Sensitivity of Embedding Strength. Showing ViT performance on 10 labels of CIFAR-10.}
% \label{AppViT_Ablation_EmbeddingStrength_cifar10_10}
%   \end{minipage}\hfill
%   \begin{minipage}{0.48\textwidth}
%     \centering
%     \includegraphics[width=\linewidth]{figs/Ablation_PN_ViT_CIFA10_Label40250_ErrorRate.pdf}
%     \caption{Sensitivity of Embedding Strength. Showing ViT performance on 40 and 250 labels of CIFAR-10.}
%     \label{AppViT_Ablation_EmbeddingStrength_cifar10_40_250}
%   \end{minipage}
% \end{figure}

% \newpage
\section{Additional Discussion}
\label{app_additional_Discussion}

\subsection{Additional Discussion on the Embedding Fusion}
\label{app_additional_discussion_EmbeddingFusion}
Despite the explanation provided in~\citep{verma2019manifold}, other perspectives might include that provided in~\citet{Li2022PN}. \citet{Li2022PN} analyse the impact of injecting noise to a learning system from information theory perspective. The author show that certain perturbations in image space help reduce the task complexity, thus enhancing the learning outcome. Extending such analysis to the embedding space might be an interesting future work.

\subsection{Additional Discussion on Heart2Heart Benchmark Results}

While our InterLUDE and InterLUDE+ are not specifically designed for open-set SSL, the performance on the Heart2Heart Benchmark is supprisingly strong. On the other hand, FlatMatch that is competitive on the classic benchmarks substantially underperform in this open-set medical imaging benchmark, we \emph{hypothesize} that this due to FlatMatch's cross-sharpness objective's goal of pulling model toward direction that is ``beneficial to generalization on unlabeled data''. The unlabeled set of TMED-2 is uncurated, containing both out-of-distribution classes as well as feature distribution shift. More study is needed to understands the challenges in this uncurated unlabeled set and the limitation of current SSL algorithms under this challenging real-world scenario.

\end{document}